%% file: main.tex
\documentclass[11pt]{article}
\usepackage[preprint]{acl}
\usepackage{times}
\usepackage{latexsym}
\usepackage[T1]{fontenc}
\usepackage[utf8]{inputenc}
\usepackage{microtype}
\usepackage{inconsolata}
\usepackage{amsmath,amssymb}
\usepackage{booktabs,tabularx,array}
\usepackage{longtable,fvextra}
\usepackage{graphicx,xcolor}
\usepackage{xurl}
\usepackage{placeins}
\usepackage[nospread]{cuted}
\usepackage{tikz}
\usetikzlibrary{arrows.meta}
\input{paper_macros}
\input{appendices/layout}
\input{tables/generated/numbers}

\title{OutfitterBench: Evaluating LLM Agents on\\
Long-Horizon Resource Allocation and Commitment Coordination}
\author{Jie Huang\textsuperscript{1} \quad
  Yuecheng Hong\textsuperscript{1} \quad
  Yao Chen\textsuperscript{2} \\
  \bfseries Shengding Jiao\textsuperscript{3} \quad
  Xudong Liu\textsuperscript{1} \quad
  Qiang Ye\textsuperscript{4} \\
  \textsuperscript{1}Harbin Institute of Technology \\
  \textsuperscript{2}Independent Researcher \quad
  \textsuperscript{3}Peking University \\
  \textsuperscript{4}University of Science and Technology of China \\
  \href{mailto:jhuang@ir.hit.edu.cn}{\texttt{jhuang@ir.hit.edu.cn}} \quad
  \href{mailto:2023112391@stu.hit.edu.cn}{\texttt{2023112391@stu.hit.edu.cn}}}

\begin{document}
\maketitle
\begin{abstract}
\input{sections/abstract}
\end{abstract}

\input{figures/value_overview}
\input{sections/introduction}
\input{sections/related_work}
\raggedbottom
\input{sections/benchmark}
\input{sections/evaluation}
\flushbottom
\input{sections/results}
\raggedbottom
\input{sections/behavioral_analysis}
\FloatBarrier
\input{sections/discussion}
\input{sections/limitations}

\bibliography{references}

\clearpage
\appendix
\input{appendices/guide}
\input{appendices/environment_details}
\FloatBarrier
\input{appendices/run_details}

\clearpage
\input{appendices/results_details}
\clearpage
\input{appendices/diagnostics}
\clearpage
\raggedbottom
\input{appendices/case_evidence}
\end{document}

%% file: paper_macros.tex
\newcommand{\bench}{OutfitterBench}
\newcommand{\decisiondays}{200}
\newcommand{\runoffdays}{14}
\newcommand{\finalday}{214}
\newcommand{\initialcash}{3{,}000}
\newcommand{\initialvalue}{3{,}397.06}
\newcommand{\capitalceiling}{2{,}141.83}
\newcommand{\indicator}{\mathbf{1}}
\newcommand{\centround}[1]{\operatorname{rnd}_{0.01}\!\left(#1\right)}
\newcommand{\code}[1]{{\small\nolinkurl{#1}}}

\newcommand{\seedresult}[3]{%
  \ifcsname #1Seed#2#3\endcsname
    \csname #1Seed#2#3\endcsname
  \else
    \PackageError{OutfitterBench}{Missing seed result #1/#2/#3}{Regenerate the paper assets.}%
  \fi}
\newcolumntype{Y}{>{\raggedright\arraybackslash}X}
\newcolumntype{P}[1]{>{\raggedright\arraybackslash}p{#1}}

%% file: appendices/layout.tex
\newenvironment{appendixfigure}{%
  \begin{strip}\begin{minipage}{\textwidth}\centering\captionsetup{type=figure}%
}{\end{minipage}\end{strip}}

%% file: tables/generated/numbers.tex
\newcommand{\StudyRunCount}{14}
\newcommand{\StudyModelCount}{7}
\newcommand{\StudyWorldCount}{2}

\newcommand{\PositiveRunCount}{10}

\newcommand{\GptNet}{+2,556.78}

\newcommand{\OpusNet}{+636.11}

\expandafter\def\csname GptSeed42Final\endcsname{5,857.44}
\expandafter\def\csname GptSeed73Final\endcsname{6,050.24}
\expandafter\def\csname GptSeed42Net\endcsname{+2,460.38}
\expandafter\def\csname GptSeed73Net\endcsname{+2,653.18}
\expandafter\def\csname GptSeed42Pickups\endcsname{263}
\expandafter\def\csname GptSeed73Pickups\endcsname{244}
\expandafter\def\csname GptSeed42Reliability\endcsname{92.93}
\expandafter\def\csname GptSeed73Reliability\endcsname{98.79}
\expandafter\def\csname GptSeed42Compensation\endcsname{426.33}
\expandafter\def\csname GptSeed73Compensation\endcsname{56.11}
\expandafter\def\csname GptSeed42Rent\endcsname{9,954.11}
\expandafter\def\csname GptSeed73Rent\endcsname{9,477.78}
\expandafter\def\csname GptSeed42Purchases\endcsname{2,392.65}
\expandafter\def\csname GptSeed73Purchases\endcsname{2,070.45}
\expandafter\def\csname GptSeed42Nominal\endcsname{95.70}
\expandafter\def\csname GptSeed73Nominal\endcsname{99.22}
\expandafter\def\csname GptSeed42Options\endcsname{90.06}
\expandafter\def\csname GptSeed73Options\endcsname{87.45}
\expandafter\def\csname GptSeed42Capital\endcsname{1.374}
\expandafter\def\csname GptSeed73Capital\endcsname{1.394}
\expandafter\def\csname GptSeed42Intrusion\endcsname{0.50}
\expandafter\def\csname GptSeed73Intrusion\endcsname{0.00}
\expandafter\def\csname GptSeed42Gaps\endcsname{8.76}
\expandafter\def\csname GptSeed73Gaps\endcsname{3.19}
\expandafter\def\csname GptSeed42NetEquipmentOutlay\endcsname{1,223.21}
\expandafter\def\csname GptSeed73NetEquipmentOutlay\endcsname{1,023.80}
\expandafter\def\csname KimiSeed42Final\endcsname{3,800.32}
\expandafter\def\csname KimiSeed73Final\endcsname{3,302.89}
\expandafter\def\csname KimiSeed42Net\endcsname{+403.26}
\expandafter\def\csname KimiSeed73Net\endcsname{-94.17}
\expandafter\def\csname KimiSeed42Pickups\endcsname{221}
\expandafter\def\csname KimiSeed73Pickups\endcsname{256}
\expandafter\def\csname KimiSeed42Reliability\endcsname{81.85}
\expandafter\def\csname KimiSeed73Reliability\endcsname{79.50}
\expandafter\def\csname KimiSeed42Compensation\endcsname{829.57}
\expandafter\def\csname KimiSeed73Compensation\endcsname{1,144.43}
\expandafter\def\csname KimiSeed42Rent\endcsname{8,022.52}
\expandafter\def\csname KimiSeed73Rent\endcsname{8,055.91}
\expandafter\def\csname KimiSeed42Purchases\endcsname{2,619.91}
\expandafter\def\csname KimiSeed73Purchases\endcsname{2,364.53}
\expandafter\def\csname KimiSeed42Nominal\endcsname{82.46}
\expandafter\def\csname KimiSeed73Nominal\endcsname{84.68}
\expandafter\def\csname KimiSeed42Options\endcsname{86.86}
\expandafter\def\csname KimiSeed73Options\endcsname{84.86}
\expandafter\def\csname KimiSeed42Capital\endcsname{0.965}
\expandafter\def\csname KimiSeed73Capital\endcsname{0.779}
\expandafter\def\csname KimiSeed42Intrusion\endcsname{3.28}
\expandafter\def\csname KimiSeed73Intrusion\endcsname{4.93}
\expandafter\def\csname KimiSeed42Gaps\endcsname{22.56}
\expandafter\def\csname KimiSeed73Gaps\endcsname{22.40}
\expandafter\def\csname KimiSeed42NetEquipmentOutlay\endcsname{1,283.25}
\expandafter\def\csname KimiSeed73NetEquipmentOutlay\endcsname{1,278.31}
\expandafter\def\csname GlmSeed42Final\endcsname{3,516.69}
\expandafter\def\csname GlmSeed73Final\endcsname{2,815.11}
\expandafter\def\csname GlmSeed42Net\endcsname{+119.63}
\expandafter\def\csname GlmSeed73Net\endcsname{-581.95}
\expandafter\def\csname GlmSeed42Pickups\endcsname{244}
\expandafter\def\csname GlmSeed73Pickups\endcsname{243}
\expandafter\def\csname GlmSeed42Reliability\endcsname{72.19}
\expandafter\def\csname GlmSeed73Reliability\endcsname{72.97}
\expandafter\def\csname GlmSeed42Compensation\endcsname{1,592.91}
\expandafter\def\csname GlmSeed73Compensation\endcsname{1,563.61}
\expandafter\def\csname GlmSeed42Rent\endcsname{8,672.66}
\expandafter\def\csname GlmSeed73Rent\endcsname{7,826.81}
\expandafter\def\csname GlmSeed42Purchases\endcsname{2,380.74}
\expandafter\def\csname GlmSeed73Purchases\endcsname{1,900.95}
\expandafter\def\csname GlmSeed42Nominal\endcsname{76.57}
\expandafter\def\csname GlmSeed73Nominal\endcsname{80.41}
\expandafter\def\csname GlmSeed42Options\endcsname{79.57}
\expandafter\def\csname GlmSeed73Options\endcsname{81.43}
\expandafter\def\csname GlmSeed42Capital\endcsname{0.892}
\expandafter\def\csname GlmSeed73Capital\endcsname{0.660}
\expandafter\def\csname GlmSeed42Intrusion\endcsname{6.28}
\expandafter\def\csname GlmSeed73Intrusion\endcsname{8.37}
\expandafter\def\csname GlmSeed42Gaps\endcsname{34.10}
\expandafter\def\csname GlmSeed73Gaps\endcsname{27.68}
\expandafter\def\csname GlmSeed42NetEquipmentOutlay\endcsname{1,256.71}
\expandafter\def\csname GlmSeed73NetEquipmentOutlay\endcsname{1,085.70}
\expandafter\def\csname DeepSeekSeed42Final\endcsname{2,752.35}
\expandafter\def\csname DeepSeekSeed73Final\endcsname{2,895.18}
\expandafter\def\csname DeepSeekSeed42Net\endcsname{-644.71}
\expandafter\def\csname DeepSeekSeed73Net\endcsname{-501.88}
\expandafter\def\csname DeepSeekSeed42Pickups\endcsname{296}
\expandafter\def\csname DeepSeekSeed73Pickups\endcsname{293}
\expandafter\def\csname DeepSeekSeed42Reliability\endcsname{74.56}
\expandafter\def\csname DeepSeekSeed73Reliability\endcsname{83.71}
\expandafter\def\csname DeepSeekSeed42Compensation\endcsname{1,699.12}
\expandafter\def\csname DeepSeekSeed73Compensation\endcsname{949.03}
\expandafter\def\csname DeepSeekSeed42Rent\endcsname{8,996.99}
\expandafter\def\csname DeepSeekSeed73Rent\endcsname{8,856.52}
\expandafter\def\csname DeepSeekSeed42Purchases\endcsname{3,810.59}
\expandafter\def\csname DeepSeekSeed73Purchases\endcsname{4,789.38}
\expandafter\def\csname DeepSeekSeed42Nominal\endcsname{68.19}
\expandafter\def\csname DeepSeekSeed73Nominal\endcsname{80.61}
\expandafter\def\csname DeepSeekSeed42Options\endcsname{92.22}
\expandafter\def\csname DeepSeekSeed73Options\endcsname{94.59}
\expandafter\def\csname DeepSeekSeed42Capital\endcsname{0.746}
\expandafter\def\csname DeepSeekSeed73Capital\endcsname{0.868}
\expandafter\def\csname DeepSeekSeed42Intrusion\endcsname{5.10}
\expandafter\def\csname DeepSeekSeed73Intrusion\endcsname{4.24}
\expandafter\def\csname DeepSeekSeed42Gaps\endcsname{40.51}
\expandafter\def\csname DeepSeekSeed73Gaps\endcsname{26.59}
\expandafter\def\csname DeepSeekSeed42NetEquipmentOutlay\endcsname{1,991.40}
\expandafter\def\csname DeepSeekSeed73NetEquipmentOutlay\endcsname{2,388.18}
\expandafter\def\csname LunaSeed42Final\endcsname{4,634.30}
\expandafter\def\csname LunaSeed73Final\endcsname{4,054.69}
\expandafter\def\csname LunaSeed42Net\endcsname{+1,237.24}
\expandafter\def\csname LunaSeed73Net\endcsname{+657.63}
\expandafter\def\csname LunaSeed42Pickups\endcsname{188}
\expandafter\def\csname LunaSeed73Pickups\endcsname{225}
\expandafter\def\csname LunaSeed42Reliability\endcsname{92.16}
\expandafter\def\csname LunaSeed73Reliability\endcsname{95.34}
\expandafter\def\csname LunaSeed42Compensation\endcsname{282.12}
\expandafter\def\csname LunaSeed73Compensation\endcsname{175.23}
\expandafter\def\csname LunaSeed42Rent\endcsname{7,836.50}
\expandafter\def\csname LunaSeed73Rent\endcsname{7,374.00}
\expandafter\def\csname LunaSeed42Purchases\endcsname{2,113.35}
\expandafter\def\csname LunaSeed73Purchases\endcsname{1,974.68}
\expandafter\def\csname LunaSeed42Nominal\endcsname{93.02}
\expandafter\def\csname LunaSeed73Nominal\endcsname{96.67}
\expandafter\def\csname LunaSeed42Options\endcsname{75.07}
\expandafter\def\csname LunaSeed73Options\endcsname{73.50}
\expandafter\def\csname LunaSeed42Capital\endcsname{1.046}
\expandafter\def\csname LunaSeed73Capital\endcsname{0.893}
\expandafter\def\csname LunaSeed42Intrusion\endcsname{2.23}
\expandafter\def\csname LunaSeed73Intrusion\endcsname{2.83}
\expandafter\def\csname LunaSeed42Gaps\endcsname{9.17}
\expandafter\def\csname LunaSeed73Gaps\endcsname{8.65}
\expandafter\def\csname LunaSeed42NetEquipmentOutlay\endcsname{1,024.41}
\expandafter\def\csname LunaSeed73NetEquipmentOutlay\endcsname{951.70}
\expandafter\def\csname MimoSeed42Final\endcsname{4,595.17}
\expandafter\def\csname MimoSeed73Final\endcsname{3,671.53}
\expandafter\def\csname MimoSeed42Net\endcsname{+1,198.11}
\expandafter\def\csname MimoSeed73Net\endcsname{+274.47}
\expandafter\def\csname MimoSeed42Pickups\endcsname{266}
\expandafter\def\csname MimoSeed73Pickups\endcsname{260}
\expandafter\def\csname MimoSeed42Reliability\endcsname{81.85}
\expandafter\def\csname MimoSeed73Reliability\endcsname{87.25}
\expandafter\def\csname MimoSeed42Compensation\endcsname{992.47}
\expandafter\def\csname MimoSeed73Compensation\endcsname{612.81}
\expandafter\def\csname MimoSeed42Rent\endcsname{9,008.24}
\expandafter\def\csname MimoSeed73Rent\endcsname{7,778.31}
\expandafter\def\csname MimoSeed42Purchases\endcsname{1,779.44}
\expandafter\def\csname MimoSeed73Purchases\endcsname{2,199.16}
\expandafter\def\csname MimoSeed42Nominal\endcsname{84.96}
\expandafter\def\csname MimoSeed73Nominal\endcsname{90.29}
\expandafter\def\csname MimoSeed42Options\endcsname{84.77}
\expandafter\def\csname MimoSeed73Options\endcsname{78.46}
\expandafter\def\csname MimoSeed42Capital\endcsname{0.999}
\expandafter\def\csname MimoSeed73Capital\endcsname{0.816}
\expandafter\def\csname MimoSeed42Intrusion\endcsname{4.13}
\expandafter\def\csname MimoSeed73Intrusion\endcsname{2.19}
\expandafter\def\csname MimoSeed42Gaps\endcsname{26.52}
\expandafter\def\csname MimoSeed73Gaps\endcsname{16.40}
\expandafter\def\csname MimoSeed42NetEquipmentOutlay\endcsname{1,002.19}
\expandafter\def\csname MimoSeed73NetEquipmentOutlay\endcsname{1,254.49}
\expandafter\def\csname OpusSeed42Final\endcsname{4,007.09}
\expandafter\def\csname OpusSeed73Final\endcsname{4,059.25}
\expandafter\def\csname OpusSeed42Net\endcsname{+610.03}
\expandafter\def\csname OpusSeed73Net\endcsname{+662.19}
\expandafter\def\csname OpusSeed42Pickups\endcsname{289}
\expandafter\def\csname OpusSeed73Pickups\endcsname{304}
\expandafter\def\csname OpusSeed42Reliability\endcsname{91.46}
\expandafter\def\csname OpusSeed73Reliability\endcsname{89.94}
\expandafter\def\csname OpusSeed42Compensation\endcsname{372.64}
\expandafter\def\csname OpusSeed73Compensation\endcsname{572.77}
\expandafter\def\csname OpusSeed42Rent\endcsname{7,496.15}
\expandafter\def\csname OpusSeed73Rent\endcsname{9,086.54}
\expandafter\def\csname OpusSeed42Purchases\endcsname{1,567.12}
\expandafter\def\csname OpusSeed73Purchases\endcsname{3,448.80}
\expandafter\def\csname OpusSeed42Nominal\endcsname{98.81}
\expandafter\def\csname OpusSeed73Nominal\endcsname{94.63}
\expandafter\def\csname OpusSeed42Options\endcsname{87.39}
\expandafter\def\csname OpusSeed73Options\endcsname{84.52}
\expandafter\def\csname OpusSeed42Capital\endcsname{0.907}
\expandafter\def\csname OpusSeed73Capital\endcsname{1.021}
\expandafter\def\csname OpusSeed42Intrusion\endcsname{1.76}
\expandafter\def\csname OpusSeed73Intrusion\endcsname{0.93}
\expandafter\def\csname OpusSeed42Gaps\endcsname{5.96}
\expandafter\def\csname OpusSeed73Gaps\endcsname{8.95}
\expandafter\def\csname OpusSeed42NetEquipmentOutlay\endcsname{834.22}
\expandafter\def\csname OpusSeed73NetEquipmentOutlay\endcsname{1,800.24}
\expandafter\def\csname CalendarSeed42Net\endcsname{+1,103.57}
\expandafter\def\csname CalendarSeed42Score\endcsname{4,500.63}
\expandafter\def\csname CalendarSeed42Pickups\endcsname{168}
\expandafter\def\csname CalendarSeed42Reliability\endcsname{93.85}
\expandafter\def\csname CalendarSeed42Compensation\endcsname{173.46}
\expandafter\def\csname CalendarSeed42Rent\endcsname{7,454.82}
\expandafter\def\csname CalendarSeed42Purchases\endcsname{1,485.36}
\expandafter\def\csname CalendarSeed42Capital\endcsname{0.967}
\expandafter\def\csname CalendarSeed42Nominal\endcsname{100.00}
\expandafter\def\csname CalendarSeed73Net\endcsname{+1,135.45}
\expandafter\def\csname CalendarSeed73Score\endcsname{4,532.51}
\expandafter\def\csname CalendarSeed73Pickups\endcsname{145}
\expandafter\def\csname CalendarSeed73Reliability\endcsname{89.51}
\expandafter\def\csname CalendarSeed73Compensation\endcsname{353.24}
\expandafter\def\csname CalendarSeed73Rent\endcsname{7,801.37}
\expandafter\def\csname CalendarSeed73Purchases\endcsname{1,574.97}
\expandafter\def\csname CalendarSeed73Capital\endcsname{0.978}
\expandafter\def\csname CalendarSeed73Nominal\endcsname{99.40}

%% file: sections/abstract.tex
In many long-horizon settings, an agent's own choices create future
constraints: committing finite capacity establishes future obligations and
forecloses later opportunities. Sustaining performance therefore requires
both keeping evolving plans feasible and comparing the value of resource
allocations with the alternatives they displace. Recent benchmarks
increasingly evaluate LLM agents in continuing operations, but give less
attention to how agents keep their own commitments feasible while
reassessing resource allocations as opportunities change.
We introduce OutfitterBench, where agents manage an
outdoor-equipment rental business for 200 days, purchasing from a
46-product catalog and handling roughly 1,100 reservation requests.
Under a fixed capital ceiling, reservations and uncertain walk-in demand
compete for reusable equipment whose availability changes over time.
Experiments with seven frontier
LLMs show that several agents underperform a simple
rule-based baseline, while only the leading agent consistently surpasses it.
Trajectory analysis shows that some agents apply known constraints
inconsistently and make new decisions that conflict with existing commitments.
Agents with more reliable fulfillment can still secure low returns from
comparable commitments or delay resource reallocation. The leading agent
better connects preparation with economic choices but also misses
opportunities for more profitable asset use. These findings highlight
limitations in sustaining coherent resource plans and reassessing their
value over time. The benchmark code is publicly available at
\url{https://github.com/adminblackjack/OutfitterBench}.

%% file: figures/value_overview.tex
\begin{figure*}[t]
\centering
\includegraphics[width=\textwidth]{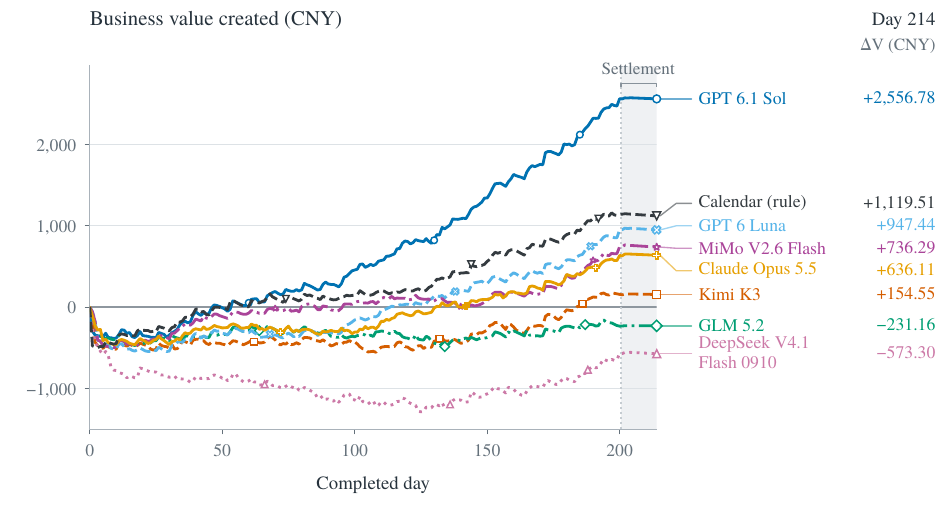}
\caption{Business value created over the season, averaged equally over
two world seeds for each agent. Curves use contract-adjusted economic
value relative to the initial shop, accounting for equipment and
outstanding obligations. Labels on the right report mean official net
value creation ($\Delta V$) at day 214. Shading marks the 14-day automatic settlement period.
Calendar provides a fixed-policy reference between the highest- and
lower-performing agents.}
\label{fig:mean-value}
\end{figure*}

%% file: sections/introduction.tex
\section{Introduction}
\label{sec:introduction}

Long-running agents often make choices that change both the resources
available for future work and the obligations they will need to meet
\citep{brehmer1992dynamic,kumar2026scarce,han2026enterprisearena}.
The value of these choices depends on how later decisions maintain or
revise the resource arrangements they create
\citep{sun2023adaplanner,you2026syncplan}. Consider a rental manager
with one tent available today. Renting it earns immediate income but may
leave it unavailable for a more valuable trip later. Holding it preserves
that opportunity, whose arrival and value remain uncertain. Once the
manager accepts a future booking, an intervening rental also affects an
existing obligation: serving both customers requires enough time for the
tent to return, be cleaned, and become ready again. Later rental decisions
must therefore weigh new opportunities against the capacity needed for
commitments already made \citep{zhang2022reusable}.

Recent business benchmarks evaluate agents through continuing operations,
including purchasing, pricing, and customer management
\citep{andonlabs2025vendingbench2,zhang2026retailbench,chen2026ceobench,zhang2026erpbench},
as well as contract selection and financing decisions with lasting
consequences \citep{he2026ycbench,han2026enterprisearena}.
Less attention has been given to how agents jointly maintain the feasibility
of their own commitments and reassess resource allocations as opportunities
change.

To address this gap, we introduce \bench{}, a benchmark in which agents
manage a simulated outdoor-equipment rental business for \decisiondays{}
days. Its design draws on work linking planning with the evaluation of
alternatives \citep{bratman1988plans} and examining how existing plans
change the value of new options \citep{horty2001options}. We focus on two
coupled demands.
\emph{Intertemporal feasibility maintenance} requires keeping changing
asset states, accepted work, and subsequent actions compatible.
\emph{Intertemporal resource valuation and reallocation} requires comparing
competing uses of equipment-time and capital when forming commitments and
revisiting existing allocations.

In \bench{}, advance reservations and uncertain walk-in demand draw from a
shared fleet whose availability changes through rental, return, cleaning,
and maintenance.
New rentals can undermine the supply supporting earlier bookings;
protecting those bookings can exclude more valuable uses. The agent chooses
bookings and prices, arranges supply, and limits walk-in use. It can
negotiate changes or cancel a booking at a compensation cost. Purchases and
sales reshape future supply; once the fixed equipment-capital ceiling
binds, new investment may require replacing existing assets. The environment
combines a product catalog derived from Taobao listings and historical
Beijing weather with models of customer demand and equipment operations.
The objective is to maximize terminal value: cash plus eligible equipment
resale value minus outstanding liabilities.

We compare \StudyModelCount{} model configurations with a public-information
rule policy on the same \StudyWorldCount{} world seeds.
Figure~\ref{fig:mean-value} summarizes the resulting value paths. The rule
policy outperforms several agents, while only the leading agent exceeds
it in both worlds. Agents can serve more bookings and still lose business
value. Some agents repeatedly schedule equipment without allowing for
turnaround; stronger runs sometimes overlook other affected commitments
when preparing supply across days. Reliable fulfillment can also coexist with
lower rents for comparable bookings, investment assessments that fail to
guide later asset choices, and protection limits that exclude usable rental
intervals. The leading agent more often connects supply preparation with
these economic comparisons, although local counterfactual tests reveal
remaining opportunities for more profitable equipment use.
Taken together, these findings reveal limits in agents' ability to keep
successive decisions compatible and to reassess resource allocations as
the value of competing uses changes.

Our contributions are threefold:
\begin{enumerate}
  \item \textbf{A benchmark for long-horizon resource allocation and
  commitment coordination.} We introduce \bench{}, a continuing rental
  environment that couples dynamic asset availability, agent-chosen future
  obligations, and uncertain opportunities under a long-term business-value
  objective.
  \item \textbf{A comparative evaluation of LLM agents.} We evaluate agents
  against a public-information rule-based baseline on matched world seeds,
  revealing substantial differences in long-term business value, with
  several agents underperforming the baseline.
  \item \textbf{A systematic analysis of intertemporal decisions.} We combine
  trajectory analysis with state-based diagnostics, matched-booking
  comparisons, and local counterfactual tests. The analysis identifies
  failures to carry constraints across decisions, weaker returns from
  comparable commitments, and gaps between investment assessments and
  subsequent resource choices.
\end{enumerate}
The benchmark code is publicly available at
\url{https://github.com/adminblackjack/OutfitterBench}.

%% file: sections/related_work.tex
\section{Related Work}
\label{sec:related}

\paragraph{Interactive and Long-Horizon Agent Benchmarks.}
Interactive benchmarks evaluate tool use, policy adherence,
and actions with observable consequences
\citep{liu2024agentbench,zhou2024webarena,yao2024taubench}.
Workplace and cross-session workflows extend these demands to coordinating
tools and using information from earlier interactions
\citep{xu2025theagentcompany,wang2025odysseybench}, while games and discovery
environments examine sustained exploration, strategic decisions, and
adaptation to evolving states or unfamiliar rules
\citep{paglieri2025balrog,qi2024civrealm,luo2025ultrahorizon}.

Continuing business simulations connect repeated operating decisions to
long-term outcomes. Vending-Bench 2, FoodTruck Bench, RetailBench, and
related e-commerce environments couple sourcing, pricing, inventory, and
cash flow
\citep{andonlabs2025vendingbench2,foodtruck2026methodology,zhang2026retailbench,shi2026merchantbench,pan2026businessarena}.
Project and enterprise simulations examine contract selection, investment,
and ongoing services
\citep{he2026ycbench,chen2026ceobench,han2026enterprisearena},
while multi-firm economies additionally study negotiation and market
interaction
\citep{sugiura2026coffeebench,zheng2026marketbench,zhang2026erpbench}.
Across these settings, earlier decisions shape the opportunities and
constraints of subsequent action.

\paragraph{Planning and Commitment Coordination.}
Maintaining a plan requires checking how actions and new observations
affect the conditions for subsequent steps \citep{ghallab2016planning}.
Planning evaluations test
plan generation, verification, and replanning
\citep{valmeekam2023planbench}. Feedback-driven methods revise plans when
observations depart from expectations \citep{sun2023adaplanner}, while
coordination frameworks explicitly manage execution dependencies and
detect stale joint plans \citep{you2026syncplan}.
Coordination can also depend on agreements formed during earlier
interactions. SCARCE studies how agents with separate objectives negotiate
access to shared execution resources and develop agreements and norms
that persist across tasks \citep{kumar2026scarce}. \bench{} examines a
different coordination problem: whether a manager can keep customer
commitments compatible with changing resource availability and subsequent
allocations.

\paragraph{Intertemporal Resource Allocation.}
A resource decision affects both current returns and what remains
available for later work. Inventory control studies examine ordering under
demand shifts and uncertain lead times \citep{baek2026inventorybench}.
When agents also choose which work to accept, resource allocation and
commitment formation become linked. In YC-Bench, accepted projects draw on
a fixed workforce, with progress accumulating toward project deadlines;
taking on additional work can overcommit shared employee effort
\citep{he2026ycbench}. CEO-Bench links growth and investment to the capacity
needed for continuing service \citep{chen2026ceobench}, while EnterpriseArena
examines how financing timing and scale affect future liquidity under
delayed settlement and changing economic conditions
\citep{han2026enterprisearena}.

Reusable-resource allocation makes the duration of occupancy part of the
economic decision: serving one customer temporarily removes capacity from
other potential uses \citep{zhang2022reusable}.
Whereas project work can accumulate toward completion before a deadline,
rental commitments require reusable capacity during specified future
service intervals. \bench{} examines how agents keep such arrangements
feasible while reassessing which resource uses to retain or revise.

%% file: sections/benchmark.tex
\section{OutfitterBench}
\label{sec:benchmark}

\bench{} simulates an outdoor-equipment rental business. Rental provides a
setting for reusable-resource allocation: accepting business commits
capacity before later demand is known
\citep{chen2017revenue,zhang2022reusable}. To study feasibility maintenance
and resource valuation together, we design the environment around three
interacting features.

First, outdoor trips require complementary equipment to be ready together,
while rental, return, and servicing change its availability. Agents must
keep accepted bookings compatible with evolving supply and subsequent uses.
Second, customers compete for shared rental capacity
\citep{savin2005capacity}: reservations and uncertain walk-ins draw on the
same fleet. Prices and capacity protection determine what commitments earn
and which opportunities remain available. Third, agents can revise
commitments and change fleet composition, but customer response delays,
cancellation compensation, and resale losses make adjustments costly.
Once the equipment-capital ceiling binds, new investment may require
replacing existing assets.

We construct the product catalog from Taobao listings and use historical
Beijing weather to model demand and operating conditions. Customer behavior
and equipment operations follow explicit rules and probabilistic models.
Figure~\ref{fig:environment} gives an overview of the environment.

\input{figures/environment_overview}

\subsection{Task, Interaction, and Objective}
\label{sec:interaction}

A season comprises \decisiondays{} manager-decision days followed by
\runoffdays{} automatic settlement days, during which outstanding rentals,
service work, and liabilities settle after new business ends. The objective
is to maximize terminal business value: cash plus eligible equipment resale
value, less outstanding liabilities. Rent generates income, while
procurement, servicing, operating expenses, and customer compensation
create costs. Section~\ref{sec:official-score} defines the score
and reported value gain.

Each morning, the environment processes deliveries, service work,
returns, and eligible pickups. The agent receives an inbox and can query
records, set prices, buy or sell equipment, arrange servicing, and handle
bookings. At closing, the environment resolves outstanding pickup issues,
serves walk-ins, charges daily costs, and advances the day. Each day starts
a fresh model session. Business records persist, and the agent can
maintain retrievable notes and a handoff for the next day.

Observations include finances, transactions, asset and contract states,
service schedules, customer feedback, and the public calendar and weather
forecast. Customer preferences are partly revealed through interactions;
future requests and operating outcomes remain uncertain.

\subsection{Customer Demand and Commitments}
\label{sec:demand}
\label{sec:reservations}

Reservations and walk-ins compete for the same fleet, linking current
admissions to capacity available for later bookings. Both channels
originate from a common pool of potential outings. For each intended
outing day \(d\) and trip type \(k\) (overnight
camping, picnic outings, or lightweight trips), the simulator samples the
number of potential customer parties:
\begin{equation}
 N_{k,d} \sim \operatorname{Poisson}(\lambda_{k,d}).
 \label{eq:potential-demand}
\end{equation}
The rate \(\lambda_{k,d}\) varies with season and calendar. Weekends and
public holidays concentrate demand; school holidays also increase weekday
activity. Each party is
assigned a channel and, for reservations, a booking lead time. Advance
customers decide whether to submit a request using the forecast available
that day for their trip; walk-ins respond to arrival-day weather. The
agent sees the same public forecast.

For each outing, we sample party composition and rental duration by trip
type and calendar, then generate required and optional equipment needs.
Public service profiles translate these needs into suitable products and
quantities using functional guarantees, such as temperature suitability,
and per-item capacities, such as sleeping places. Budget and preference
parameters come from predefined scenario distributions, allowing parties
with similar needs to value equipment and rental convenience differently.
Equipment choices within a reservation share budget and price-sensitivity
parameters, with preferences also varying by category.

For advance inquiries, we use a random-utility model
\citep{train2009discrete} to compare complete rental plans with buying
equipment, using gear already available to the party, or abandoning the
trip. A product's functional fit and quality contribute to both purchase
and rental utility \citep{lancaster1966consumer}. Ownership benefits,
service convenience, payments, and the rental deposit then affect the
relative appeal of the alternatives. A quoted plan is acceptable when its
utility exceeds that of the outside alternatives. These alternatives
therefore constrain what customers are willing to pay for a rental plan.
Walk-ins choose required equipment first; optional choices depend on
those selections.

Advance requests specify equipment requirements, trip lengths, and
permitted pickup dates. We include two forms to examine how agents choose
which work to undertake and on what terms. Fixed-term offers provide
customer-approved configurations and rents; the agent selects a plan and
date or declines. For manager-priced inquiries, it proposes configurations
and prices for customer approval. Equipment choices and prices shape both
the return from a commitment and the capacity left for other opportunities,
linking pricing with resource allocation
(\mbox{\citealp{bitran2003pricing}};
\mbox{\citealp{talluri2004theory}}).
Customer approval can directly confirm a quoted booking.
Confirmation fixes products, quantities, dates, and prices as obligations
that the agent must support or revise through later decisions.

Walk-ins are served automatically at closing using posted prices and
ready equipment. The agent controls new walk-in occupancy through product-
and date-specific limits on how many units those rentals can occupy.
Walk-in rentals can extend into future booking windows, and new
reservations can overlap earlier bookings. Through both channels, the
agent commits capacity before knowing which later requests will arrive
or what they will be worth.

To support learning from experience, the environment provides transaction
histories, unfilled walk-in requests, and occasional noisy feedback on
declined quotations, including concerns about price or equipment fit.
Together with public product attributes, these records give agents evidence
about how customer responses vary with equipment, prices, and trip
conditions.

\subsection{Shared Capacity and Resource Allocation}
\label{sec:assets}
\label{sec:revision}

Meeting accepted commitments requires coordinating complementary assets
whose readiness depends on prior rentals and shared turnaround capacity.
The shop starts with 27 physical assets and can purchase from a 46-product
catalog of tents, sleeping equipment, and picnic supplies.
Different parties can draw on overlapping sets of suitable products,
while their functional requirements limit substitution. A booking may
require several equipment types together, so a shortage can prevent
complete fulfillment even when other equipment is idle. Confirmed products
remain binding unless the customer authorizes a change.

\input{figures/equipment_tradeoffs}

Each unit has its own condition, usage history, and operational status.
Once rented, a unit becomes available again only after returning and
completing required cleaning, repair, or maintenance. Returns enter a
shared, capacity-limited cleaning queue and become available no earlier
than the next day. Cleaning workload determines both the cleaning charge
and the demand placed on the shared queue. The shared queue can delay
readiness for other orders, even across equipment types or non-overlapping
rental intervals.

Calendar and weather effects connect demand with these constraints.
Holidays can bring more and longer trips while making renters more likely
to extend their use; weather exposure changes cleaning workloads. Updated
forecasts can prompt rescheduling or cancellation. Public information and
operating history help anticipate these patterns, although
individual outcomes remain uncertain.

The agent can accept bookings against expected deliveries, returns, and
service completions, with units assigned at pickup. Confirmation records
the agreed obligations without checking future capacity; the agent must
arrange supply and keep later allocations compatible with existing bookings.
Walk-in limits can protect bookings but also exclude rentals that would
leave equipment ready again before the booked pickup.
Figure~\ref{fig:equipment-tradeoffs} illustrates these competing uses.

As conditions change, the agent can reassess how and whether to support a
commitment. Before pickup, it can propose revisions; customers respond the
next day, and accepted terms take effect
when the agent applies them. Substitutions or reduced delivery at pickup
require authorization. Alternatively, the agent can cancel, refund
unearned payments, and pay compensation that increases as pickup
approaches. Waiting may improve information but leave fewer adjustment
options. The agent compares the value of fulfillment with that of
releasing capacity, accounting for agreed rent, further preparation, and
the cost of changing course.

Fleet investments compete for a fixed equipment-capital allowance.
Purchases require cash and must satisfy:
\begin{equation}
 B_t + I_t \leq K_{\max},
 \label{eq:equipment-capital}
\end{equation}
where \(B_t\) is the historical acquisition cost still counted against
the ceiling, and \(I_t\) is the cost of accepted inbound orders.
Sale or disposal releases capital basis the next morning; until then,
that basis remains in \(B_t\). Revenue adds cash without raising
\(K_{\max}\). Owned equipment also incurs daily holding charges.
Dealer resale value starts at 55\% of purchase cost for intact, unused
equipment and declines with condition and wear.
When the ceiling binds, new purchases may require replacing
existing assets. The agent compares the future uses of current and
replacement assets, accounting for resale losses, holding costs, delivery
waits, and existing booking obligations.

\subsection{Data Sources and Parameterization}
\label{sec:construction}

We use DeepSeek V4 Flash 0731 as an auxiliary model to clean and structure
product information from Taobao listings. Specifications provide functional
attributes and per-item capacities; listing prices inform procurement
costs and customers' purchase alternatives, with selected adjustments
recorded as benchmark settings. Products are matched to service profiles
through rules on these attributes.
The model also estimates product-level durability and cleaning workload
from listed materials and construction details. Shared
operating and cost rules translate these priors into wear accumulated
through rental days and turnovers, cleaning-queue workloads, and service
charges. Before freezing the data, we apply deterministic checks to verify
profile membership against product specifications, validate capacities,
and confirm a compatible profile for each rental product.

Customer budget and preference distributions, together with behavioral
response parameters, are specified as benchmark settings.

Historical Beijing weather is replayed by month and day. We synthesize
public forecasts by perturbing temperatures and weather categories using
lead-time-dependent error settings. The demand and operating models use
this shared weather sequence and the public calendar.

Random draws are indexed by event and world seed to support reproducible
simulation.
Appendices~\ref{app:environment} and~\ref{app:runs} provide operating rules
and the evaluated configuration.

%% file: figures/environment_overview.tex
\begin{figure*}[!t]
\centering
\includegraphics[width=\textwidth]{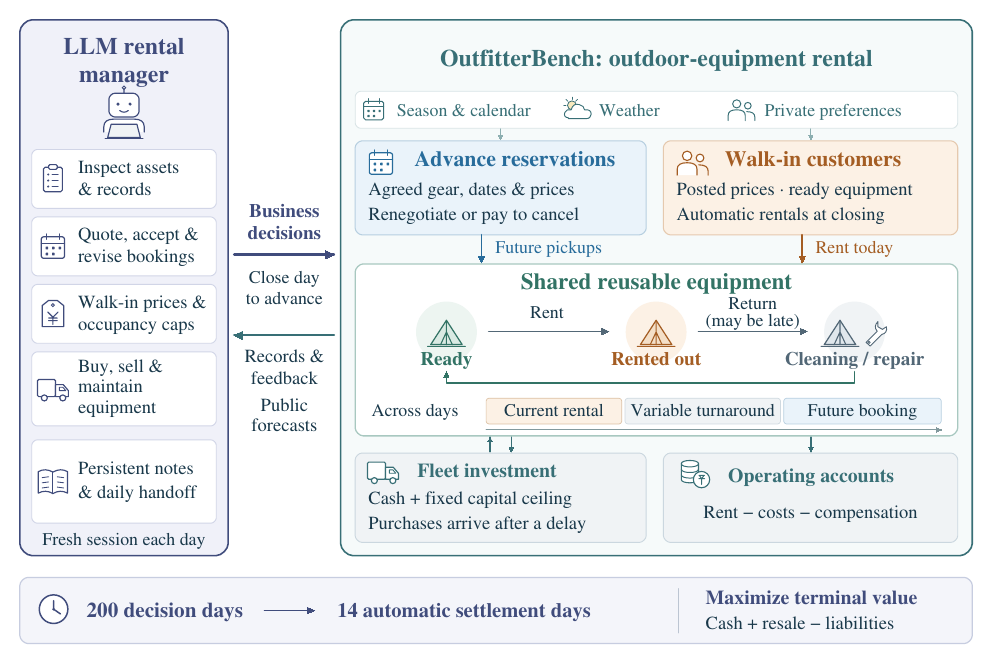}
\caption{The \bench{} rental shop combines two demand channels with a
reusable equipment fleet. The agent manages bookings, walk-in prices and
occupancy limits, and equipment investment using business records, public
forecasts, and customer feedback. Calendar and weather jointly shape
demand and the timing of equipment reuse, while a fixed equipment-capital
ceiling constrains fleet expansion. Performance after \decisiondays{}
decision days and
\runoffdays{} automatic settlement days is terminal cash plus eligible
equipment resale value, less outstanding liabilities.}
\label{fig:environment}
\end{figure*}

%% file: figures/equipment_tradeoffs.tex
\begin{figure*}[!t]
\centering
\includegraphics[width=\textwidth]{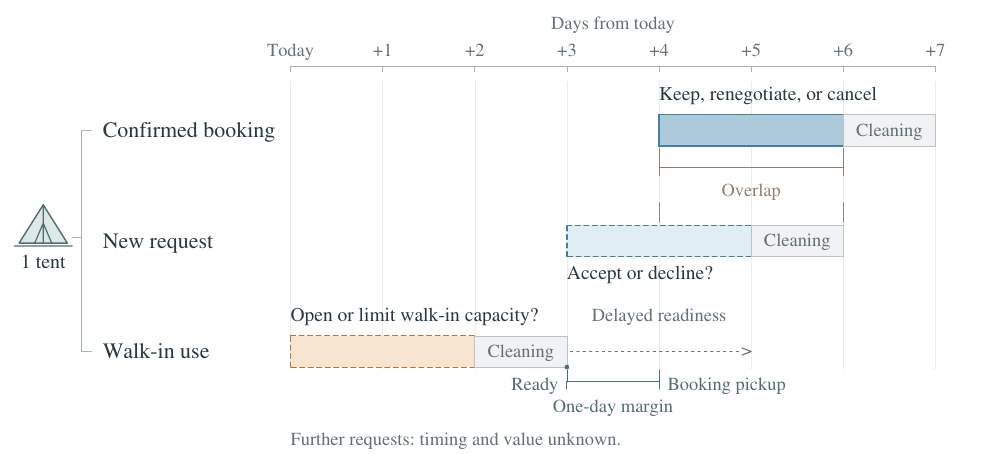}
\caption{Competing uses of a single tent. The new request overlaps a
confirmed booking, whereas a walk-in rental could fit before it with a
one-day margin after cleaning. Late return or servicing delays could
eliminate this margin. Solid and dashed outlines denote confirmed and
candidate rentals, respectively.}
\label{fig:equipment-tradeoffs}
\end{figure*}

%% file: sections/evaluation.tex
\section{Evaluation Protocol}
\label{sec:evaluation}

\subsection{Models and Operating Protocol}
\label{sec:execution-protocol}

We evaluate \StudyModelCount{} model configurations on
\StudyWorldCount{} world seeds, with one complete run per model--seed pair.
We use GPT 6.1 Sol (Sol) and Claude Opus 5.5 (Opus) with high reasoning
effort, and Kimi K3 (Kimi), GLM 5.2 (GLM), DeepSeek V4.1 Flash 0910
(DeepSeek), GPT 6 Luna (Luna), and MiMo V2.6 Flash (MiMo) with max
reasoning effort.

Each run starts with \initialcash{} CNY, the same initial fleet,
a \capitalceiling{} CNY equipment-capital ceiling, and a 15 CNY daily
operating fee. Each evaluated season generates approximately 1,100
reservation requests. Agents receive the same business tools and instructions,
fresh daily sessions, persistent records, and up to 200 model responses
per day. Run and controller details appear in Appendix~\ref{app:runs}.

Runs within each seed share the initial shop
and seeded external processes; their decisions then produce different
fleets, obligations, and customer interactions. The main results table,
value paths in Figure~\ref{fig:mean-value}, and main-text process diagnostics
average seed-level measurements equally across these worlds.
Appendix~\ref{app:results} reports individual outcomes for each model--seed pair.

\subsection{Public-Information Rule Reference}
\label{sec:evidence}

We also run Calendar, a fixed operating policy, on both worlds.
It uses a twelve-product assortment with fixed prices, replenishment
targets, maintenance, and recovery rules. Using public
asset states, return estimates, and service schedules, it accepts
bookings with positive estimated direct contribution and sufficient
projected calendar capacity.

Calendar allows one day for ordinary turnaround and interacts through
agent-visible tools. It provides a reference for the value of
consistent routines. Appendix~\ref{app:rule-baselines}
gives its operating rules.

\subsection{Operating Outcomes}
\label{sec:official-score}

At terminal day \(T=\finalday\), the score of a solvent shop is
\begin{equation}
 S_T=C_T+L_T-\Lambda_T,\qquad \Delta V=S_T-V_0,
 \label{eq:score}
\end{equation}
where \(C_T\) is cash, \(L_T\) eligible equipment resale value, and
\(\Lambda_T\) outstanding liabilities, including refundable deposits
and unearned reservation payments. We report value creation \(\Delta V\)
relative to the initial business value \(V_0=\initialvalue\) CNY.
Insolvency ends a run with a failure score. Model inference costs are recorded
separately from business value; valuation details are in
Appendix~\ref{app:valuation}.

Reservation deliveries and reliability describe the business served.
Reliability is deliveries divided by deliveries plus merchant-caused
failures; ordinary customer cancellations are excluded.

\subsection{Process Diagnostics}
\label{sec:diagnostics}

We use four diagnostic families to relate operating outcomes to feasibility
maintenance and resource valuation. All are computed from recorded states
and actions without an LLM judge. Figure~\ref{fig:behavioral-overview}
reports equal-weight means across seeds, with ratios calculated within
each run.

\paragraph{Pending-booking gaps.}
After each day's walk-ins, we check whether pending reservations have
sufficient projected supply, using public asset states, service schedules,
and return and delivery estimates. We report the share of observed
booking--days with a shortage, both overall and one or two days before
pickup. These rates track whether accepted bookings remain supported over time.

\paragraph{Action-induced new gaps.}
We compare projected coverage immediately before and after a supported
action, counting calls that leave a previously covered booking short.
Rates are calculated separately for each action type over relevant
successful calls. Quotation calls can directly confirm bookings after
customer approval, so they can also create projected gaps for existing
commitments. Walk-in rentals use customer-trip batches instead.
This check identifies subsequent choices that undermine the supply
supporting existing commitments.

\paragraph{Capital productivity.}
We measure net service contribution per 100 capital-CNY-days. Contribution
includes rent and damage deductions, less service costs and estimated
equipment-use consumption; the denominator accumulates occupied equipment
capital, including inbound purchases. This describes the return generated
by the resources tied up over time, complementing the official score's
measure of total business value.

\paragraph{Unrealized gains from capacity relaxation.}
At 40 fixed dates per run, we test whether slightly relaxing one product's
walk-in limit improves estimated contribution under a conservative supply
check. The test holds stock, prices, commitments, and that day's customer
demand fixed. We report how often a positive estimated gain remains available
under existing limits, and its mean across all sampled days, including
zero-gain dates, using a three-day turnaround allowance. Candidates depend
on each run's fleet and existing limits, so the results describe remaining
opportunities within that configuration. We therefore interpret these local
estimates alongside the booking-gap rates.

Full definitions, individual results, and counts appear together in
Appendix~\ref{app:diagnostics}, following the same four-family order.
Trajectory analysis connects these measurements to recorded information,
executed choices, and subsequent outcomes. Matched bookings and selected
historical-state probes provide further comparisons, documented in
Appendix~\ref{app:cases}.

%% file: sections/results.tex
\section{Results and Analysis}
\label{sec:results}
\label{sec:behavior}

In this section, we compare agents' long-term value creation and examine
the resource-management patterns behind their performance. We focus on
how agents maintain feasible plans and choose or revise resource
allocations as opportunities change.

\subsection{Overall Performance}
\label{sec:overall-performance}

All \StudyRunCount{} LLM runs complete the season, and \PositiveRunCount{}
create positive net value. Table~\ref{tab:main-results} compares agents and
the rule reference using equal-weight means over the two worlds.
Individual results for each world are reported in
Appendix~\ref{app:world-outcomes}.
The averaged paths in Figure~\ref{fig:mean-value} show the scale of the
resulting differences; individual paths appear in
Appendix~\ref{app:financial-paths}.

\input{tables/generated/main_results}

\paragraph{A fixed policy provides a substantial operating reference.}
Calendar creates an average of about 1,120 CNY, outperforming
Kimi, GLM, DeepSeek, and Opus on both seeds. Sol exceeds Calendar in
each world, with mean value creation of \GptNet{} CNY.
Luna and MiMo occupy the middle:
they exceed Calendar in one world and fall below it in the other.
Opus creates an average of \OpusNet{} CNY, remaining profitable in both worlds
without reaching the rule reference.
These results distinguish profitable routine operation from the gains
achievable through more effective management.

The scale of the operation does not determine this ordering. DeepSeek and
Opus each deliver more bookings than Sol in both worlds. Calendar serves
fewer reservations and earns 56--61\% of its rental income from walk-ins.
Its total rent is lower than Kimi's, GLM's, DeepSeek's, and Opus's in each
world, but lower costs more than offset the difference. The decomposition
in Appendix~\ref{app:financial-decomposition} reconciles revenue, actual
expenses, and recovered equipment value with the final score.

Schedule reliability alone does not explain the value differences. Sol,
Opus, and Luna maintain relatively reliable schedules, but Sol creates
substantially more value. We use the diagnostics in
Figure~\ref{fig:behavioral-overview} and recurring decisions in the
trajectories to examine how agents maintain feasible resource plans and
reassess the value of competing uses.

%% file: tables/generated/main_results.tex
\begin{table*}[tb]
\centering
\small
\setlength{\tabcolsep}{4pt}
\begin{tabular}{@{}lrrrr@{}}
\toprule
Agent & $\Delta V$ (CNY) & Bookings & Reliab. (\%) & Compens. (CNY) \\
\midrule
GPT 6.1 Sol & +2,556.78 & 253.5 & 95.86 & 241.22 \\
GPT 6 Luna & +947.44 & 206.5 & 93.75 & 228.68 \\
MiMo V2.6 Flash & +736.29 & 263.0 & 84.55 & 802.64 \\
Claude Opus 5.5 & +636.11 & 296.5 & 90.70 & 472.71 \\
Kimi K3 & +154.55 & 238.5 & 80.68 & 987.00 \\
GLM 5.2 & -231.16 & 243.5 & 72.58 & 1,578.26 \\
DeepSeek V4.1 Flash 0910 & -573.30 & 294.5 & 79.14 & 1,324.08 \\
Calendar (rule) & +1,119.51 & 156.5 & 91.68 & 263.35 \\
\bottomrule
\end{tabular}
\caption{Full-season outcomes averaged equally across the two world seeds. $\Delta V$ is value created above the initial business value; bookings are reservation pickups, including agreed reduced delivery. Reliability is the mean of the two run-level rates. Appendix~\ref{app:world-outcomes} reports the individual outcomes for each world.}
\label{tab:main-results}
\end{table*}

%% file: sections/behavioral_analysis.tex
\begingroup
\raggedbottom
\widowpenalty=10000
\clubpenalty=10000

\input{figures/behavioral_overview}

\subsection{Intertemporal Feasibility Maintenance}
\label{sec:finding-continuity}

Figure~\ref{fig:behavioral-overview}A shows mean projected gap rates of
roughly 6--9\% of pending booking--days for Sol, Opus, and Luna, compared
with 21--34\% for DeepSeek, Kimi, GLM, and MiMo. These groups remain
separated in each world. The trajectories reveal failures to apply
operating rules consistently and reconcile new plans with existing
obligations.

\paragraph{Applying operating rules consistently.}

Because returned equipment must be cleaned before reuse, feasible schedules
need turnaround room between consecutive rentals. Using each day's closing
equipment and booking ledger, we check whether newly confirmed bookings can
be covered when units can be reused on their return day or instead require
one day of turnaround. Requiring that extra day lowers coverage by 4.9--21.5
percentage points for DeepSeek, Kimi, GLM, and MiMo, compared with 0--3.7
points for Sol, Opus, and Luna. More of the former group's accepted
schedules therefore depend on tightly spaced uses of the same equipment.
Appendix~\ref{app:calendar-sensitivity} gives the nominal-calendar assumptions and
per-run results.

DeepSeek's records show how an unstable working rule contributes to this
pattern. Across its two runs, tables, pads, and chairs are repeatedly planned
for their return date, despite separate correct applications of the cleaning
requirement. Early corrections and cancellations do not make the rule govern
later decisions consistently. In a late-season chair review, DeepSeek first
recognizes that only one chair will be usable against three required. Seeing
that the pickup is already confirmed for the chairs' return day, it takes
that existing confirmation as evidence that return-day reuse is allowed.
Confirmation does not certify future supply, but DeepSeek records this
interpretation and carries it into the next day. The chairs return on time
but remain in cleaning at pickup. The failure follows a change to a working
rule that DeepSeek had correctly applied in earlier decisions.

\paragraph{Coordinating new plans with existing commitments.}

Across both runs, Sol repeatedly relates the quantities needed for upcoming
bookings to the dates on which equipment is expected to become usable.
These projections distinguish ready units from supply awaiting returns,
cleaning, or delivery, and guide procurement, asset retention, and walk-in
limits.

For a revised cookware schedule, Sol identifies a shortfall because a
returning unit still needs cleaning before reuse. Cash is sufficient, but
equipment capital is tied up. It sells a worn mat and buys another cookware
unit after the capital allowance is released the next day. The order arrives
two days later than estimated but before pickup. Before other asset sales,
Sol checks remaining stock against near-term bookings. As returns and
cleaning alter projected supply, it revises walk-in limits.

Maintaining coverage also requires checking whether new actions displace
supply already promised elsewhere. Figure~\ref{fig:behavioral-overview}B shows that,
on average, roughly 2--5\% of relevant confirmation calls by Sol, Opus, and
Luna introduce a new projected gap for a previously accepted booking,
compared with 14--20\% for DeepSeek, Kimi, GLM, and MiMo. This separation
holds in each world and recurs in the mean rates for quotation, fixed-offer
acceptance, and walk-in exposure.

Sol nevertheless double-books its only long-runtime lantern. After querying
an existing booking, it accepts another with overlapping rental dates and
an earlier pickup, describing the offer as supportable. The new customer
collects the lantern first; the booking accepted earlier is cancelled after
renegotiation fails. Both obligations were visible when the new one was
accepted, and the overlap does not depend on a return or cleaning forecast.

Opus correctly uses a cleaning-queue check to retain a table-and-chair
pickup. On a later queue review, however, it prepares for a chair shortage
several days away while overlooking sleeping bags and pads that will not
be ready for the next morning's booking. Both agents can organize supply
for the task at hand. Their checks sometimes omit visible constraints
relevant to other commitments. Appendix~\ref{app:continuity-cases}
documents these preparation and joint-feasibility sequences.

\subsection{Intertemporal Resource Valuation and Reallocation}
\label{sec:finding-value}

Figure~\ref{fig:behavioral-overview}C reports mean
capital productivity of CNY 1.38 per 100 capital-CNY-days for Sol, compared
with 0.96--0.97 for Opus and Luna, with comparable capital occupancy.
We examine how agents value commitments, carry feedback and prior
evaluations into resource reallocation, and use available capacity while
protecting future obligations.

\paragraph{Choosing which commitments to make and retain.}

Accepting a commitment ties up future capacity for an agreed return and
limits which later opportunities can be pursued.
Opus completes more bookings than Sol in both worlds without attaining
comparable business value. Among 77 pairs of fully delivered bookings
matched by customer request, equipment, quantities, rental dates, and
promised service, it charges less in 75; total agreed rent on these pairs
is 30--37\% lower than Sol's. Its pricing records repeatedly recommend
concessions after rejected quotes.

Those commitments also shape later admission decisions. Two four-day
table, chair, and cookware bookings earn CNY 33 each for Opus and CNY 60
for Sol; Opus later sets aside other offers because the cookware is
committed. Its larger completed operation is therefore accompanied by
weaker value capture from the equipment-time it commits.
Appendix~\ref{app:matched-bookings} gives the paired transactions and pricing
records; Appendix~\ref{app:financial-decomposition} provides the full-run
accounting.

Once a concrete conflict brings competing obligations into view, however,
even agents with weaker overall returns or unstable schedules can weigh exit
costs against the value of retaining a booking. Facing two bookings that
compete for a sleeping bag and pad, DeepSeek compares their rents,
cancellation charges, and occupied time, then pays CNY 7.86 compensation
before pickup to cancel the CNY 13.20 booking and preserve the CNY 73.50
booking, which is subsequently fulfilled. Similarly, when a tent conflict
cannot be resolved by rescheduling, Opus compares buying another tent
against early cancellation and exits the booking. The larger difference
lies in how consistently such comparisons shape commitments and resource
allocations throughout the season. Both decisions are documented under
``Deliberate exit after economic comparison'' in
Appendix~\ref{app:deliberate-exit}.

\paragraph{Turning feedback and evaluations into resource reallocation.}
\label{sec:finding-reallocation}

When resources are fully committed, acting on new opportunities requires
revisiting existing allocations and previously evaluated alternatives.
Sol and Luna can both assess alternative investments, but differ in how
those assessments shape later resource decisions. In one shared
world, Luna already holds a higher-cost comfort-chair model (about CNY 103
per unit) when it evaluates a lower-cost comfort chair (about CNY 36) for
expansion. It considers the candidate's payback, delivery time, and an asset
sale that could fund the purchase. About two weeks later, with nearly all
capital committed, it reviews its higher-cost existing chair and supports
retention through past rental income and continued earning potential. Luna
mentions the lower-cost model as a product it does not currently own,
without revisiting the earlier funding proposal. Luna continues buying and selling other
equipment, but purchases the lower-cost comfort chairs more than two months
after first assessing them.

Sol, by contrast, repeatedly carries demand comparisons into allocation
decisions. After reviewing recent demand, fleet sizes, delivery time, and
holding costs, it buys a sixth comfort chair rather than a seventeenth
lightweight chair. The lightweight fleet has earned more total rent,
but the additional comfort chair would enable six-person comfort-seating
requests. Customer rejections of quoted lightweight-chair configurations
prompt another adjustment: Sol checks that the remaining lightweight
chairs cover near-term bookings, sells an unused chair, and buys another
comfort chair once capital is released. As cold-weather demand weakens,
Sol shifts from sleeping equipment to seating.

Sol more consistently carries investment assessments into later decisions
about which assets to retain and replace. Between Luna's initial assessment
and eventual purchase, the shops receive walk-in requests for the same
141 comfort-chair units; Sol supplies 46 and Luna 16 under their respective
fleets, prices, and capacity limits. Appendix~\ref{app:capital-comparisons}
documents the decision sequences and the scope of the walk-in and
reservation comparisons.

\paragraph{Using available capacity while protecting commitments.}

Future commitments require capacity at particular times, leaving
intervening availability that may support additional work.
In the MiMo and DeepSeek replays, many additional rentals would occupy
equipment needed for upcoming bookings, leaving fewer opportunities that
fit around the existing schedule.
Figure~\ref{fig:behavioral-overview}D finds positive
estimated contribution from a small limit relaxation without a new
projected booking gap on an average of 31.25\% of sampled days for Luna,
compared with 12.50\% for Sol. For Luna, they recur across products
early and late in both seasons.

A sleeping-pad booking illustrates one way those opportunities are excluded.
Luna blocks its ready three-person pad from walk-ins twelve days before
pickup and keeps the restriction through the rental. In our capacity-relaxation
test using the following day's state, raising the limit by one unit admits
an intervening rental scheduled to return a week before pickup, with rent
exceeding its estimated cleaning and usage costs. The restriction protects a
particular future use by closing a much longer available interval. The
case highlights the value of matching protection to the timing and
capacity required by a commitment.
Unused opportunities also remain in both Sol runs.
Appendix~\ref{app:capacity-cases} reports the rental intervals and local test
results.

\medskip
\begin{samepage}
\noindent Effective preparation for individual commitments does not
consistently extend to joint feasibility, while more reliable fulfillment
can coexist with weak returns and delayed resource reallocation. Sol more
effectively compares competing uses of equipment and capital, and carries
these comparisons into supply preparation and fleet changes.
Its scheduling conflicts and unused rental
opportunities expose limits in both feasibility maintenance and value
comparison, even in the strongest runs.
\par
\end{samepage}

\endgroup

%% file: figures/behavioral_overview.tex
\begin{figure*}[t]
\centering
\includegraphics[width=\textwidth]{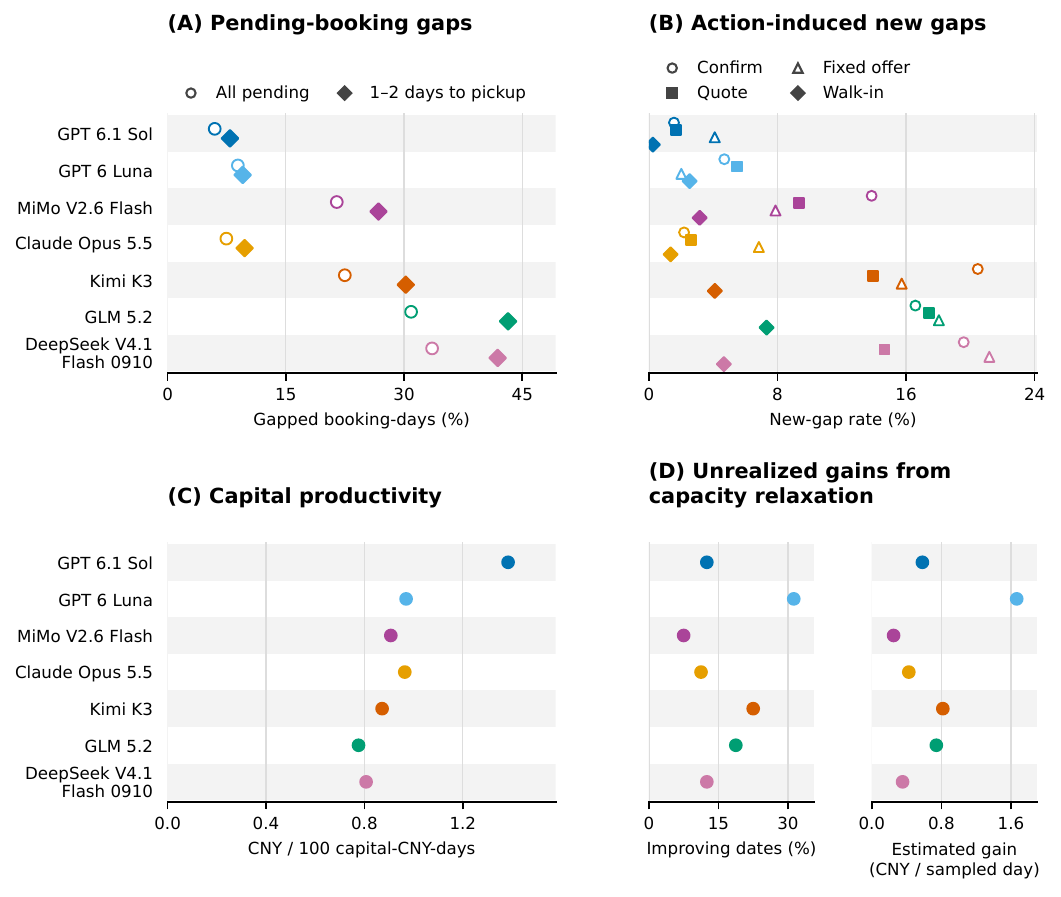}
\caption{Diagnostics for LLM agents, averaged equally across two
seeds. A reports projected gaps in pending bookings; B measures new gaps
introduced into existing bookings by subsequent actions and walk-in
rentals. C measures net service contribution per occupied capital-time.
D summarizes estimated gains from relaxing walk-in limits under a
conservative capacity check with a three-day turnaround allowance. It
reports the share of improving dates and the mean gain across all sampled
dates, including zeros. These opportunities depend on each run's fleet and
limits. Section~\ref{sec:diagnostics} defines the diagnostics;
Appendix~\ref{app:diagnostics} gives calculation details and results by model
and seed.}
\label{fig:behavioral-overview}
\end{figure*}

%% file: sections/discussion.tex
\section{Discussion and Conclusion}
\label{sec:discussion}
\label{sec:conclusion}

We introduced \bench{} to study how agents keep their own commitments
feasible while reassessing the value of resource allocations as
opportunities change.
Calendar outperforms several LLM agents, while only the leading agent
exceeds this fixed reference in both evaluated worlds.
The trajectories reveal weaknesses both in sustaining feasible arrangements
and in choosing worthwhile resource uses. Constraints correctly applied
during preparation are sometimes omitted from later decisions.
More reliable runs can still earn less from comparable bookings or delay
capital reallocation.

Feasibility maintenance and resource valuation are coupled. Existing
commitments restrict subsequent uses, while alternative returns and
adjustment costs guide which arrangements are worth retaining or revising.
Paying to exit a lower-value booking can preserve more valuable business,
whereas replacing equipment can remove support for an existing obligation.
Capacity-relaxation tests illustrate a further trade-off: more precise
protection can admit intervening rentals with positive estimated
contribution while preserving projected booking coverage. Long-horizon
consistency thus requires stable use of operating constraints and
selective revision of commitments and allocations.

These findings motivate testing decision support that combines
dependable constraint checks with selective economic reassessment. Such
support could link proposed changes to affected commitments and retain
evaluated alternatives with the conditions that would make them worth
revisiting. \bench{}
provides a testbed for studying how agents maintain feasible plans
while choosing and revising resource uses for long-term value.

%% file: sections/limitations.tex
\section{Limitations}
\label{sec:limitations}

The environment draws on real product and weather records, but customer
behavior is rule-based rather than LLM-driven, and costs and operating
events follow simulation assumptions. Transfer to other markets or real
operations remains untested. With only two world seeds and one run per
model--seed pair, the stability of the observed rankings under broader
sampling remains uncertain. Diagnostics and selected
cases help interpret decisions without isolating each behavior's
contribution to the final score; capacity-relaxation tests do not follow
subsequent agent decisions and thus do not estimate additional seasonal
profit. We use fixed business tools and a
daily-session controller, without a general-purpose shell such as Bash
or coding-agent harnesses such as Codex or Claude Code.

%% file: appendices/guide.tex
\section*{Guide to the Appendices}

Appendix~\ref{app:environment} gives operating and scoring rules;
Appendix~\ref{app:runs} describes the evaluation protocol, agent instructions
and tools, and Calendar.
For the individual outcomes behind Table~\ref{tab:main-results} and
Figure~\ref{fig:mean-value}, see Appendix~\ref{app:results}.
Appendix~\ref{app:diagnostics} follows the four diagnostic families in
Figure~\ref{fig:behavioral-overview}, with definitions and detailed results
together. Appendix~\ref{app:cases} documents the decision sequences
discussed in Section~\ref{sec:behavior}.

%% file: appendices/environment_details.tex
\section{Operating Rules and Scoring}
\label{app:environment}

This appendix specifies the operating rules behind the decisions discussed
in Sections~\ref{sec:benchmark} and~\ref{sec:behavior}.
The evaluated parameter settings appear in Appendix~\ref{app:study-config}.

\subsection{Asset Readiness and Servicing}
\label{app:units}
\label{app:day-sequence}
\label{app:service}

Service Profiles specify functional guarantees and capacity per physical
asset. The quotation workspace translates customer needs into quantities.
Rental occupancy runs from pickup \(p\) to scheduled morning return \(r\),
over \([p,r)\), and may extend if the customer returns late.

Morning processing handles deliveries, service work, returns, and eligible
pickups. Existing cleaning jobs receive capacity before newly returned
equipment joins the queue. Cleaning processes 30 work points each morning
at 0.50 CNY per point; workload varies with equipment and use.
Repair and maintenance require cash and downtime while cumulative wear
continues to constrain service life.

\subsection{Booking Formation and Revision}
\label{app:quoting}
\label{app:tools}

Advance requests are manager-priced inquiries or fixed service offers,
with one or two advertised dates. A quoted menu applies to both dates;
the manager can select one authorized complete plan and one date.
A single authorized plan with a single date confirms immediately.
Other unresolved confirmations default at closing to the original date
and customer-preferred plan. Fixed offers are accepted as whole plans.

Each line selects one product/Profile pair, with asset quantity equal
to the ceiling of service units divided by capacity per asset.
Confirmation locks products, quantities, dates, prices, and deposits;
physical assets are assigned at pickup. A 25\% prepayment is held as
unearned revenue until pickup. All selected lines, including optional
items, become obligations. Resource coverage is checked by the manager.
Additional rental days are discounted relative to the first day.

\paragraph{Protecting and revising bookings.}
\label{app:recovery}
Walk-in limits cap new occupancy by SKU and date. Existing rentals
continue, and the manager must allow for return and cleaning when
setting these limits. Proactive renegotiation proposes concrete changes
at least two days before pickup; the permitted inquiry returns the
next day, and an accepted change requires explicit application that day.
Pickup-time recovery separately obtains current customer authorization
for substitutions or reduced delivery, with delivery checked at closing.
The manager evaluates how either change affects other bookings.

\subsection{Equipment Capital, Exit, and Trials}
\label{app:capital}

Let \(K_0\) be the initial fleet's historical basis and \(H\) additional
headroom. The fixed ceiling is \(K_{\max}=K_0+H=\capitalceiling\).
With \(B_t\) owned basis still charged and \(I_t\) accepted inbound basis,
available capital is
\begin{equation}
 K_t^{\mathrm{free}}=K_{\max}-B_t-I_t.
 \label{eq:headroom}
\end{equation}
Purchases also require cash and warehouse space. Delivery transfers
basis from \(I_t\) to \(B_t\). Sale proceeds arrive immediately, while
sale or disposal releases historical basis the following day;
pending releases remain in \(B_t\) until then. The ceiling remains
fixed as the business earns revenue. The warehouse limit is
5,000 units, making capital the tighter fleet-expansion constraint.

Owned equipment incurs a daily capital holding charge of \(0.12P_a/365\),
where \(P_a\) is its acquisition basis, plus on-site non-rented storage
charges.

\paragraph{Supplier trials.}
\label{app:trials}
Temporary supplier trials provide another source of equipment:
at most four accepted trials per season and one active at a time,
with handling and usage fees, a fixed return window, and a three-rental
limit. Returning a trial item removes it from service.

\subsection{Compensation and Refunds}
\label{app:guarantee}

Each promised line receives a first-day guarantee basis \(G_i\) at
confirmation: first-day unit tariff times quantity for ordinary quotes,
or the whole-line first-day amount for fixed offers.
For affected original obligations \(A\) and \(N\) days' notice,
\begin{align}
 \alpha(N)&=1-0.75\min\!\left(\frac{\max(N,0)}7,1\right),\\
 C_A(N)&=\centround{\alpha(N)\sum_{i\in A}G_i}.
 \label{eq:compensation}
\end{align}
At least seven days' notice costs 25\% of the basis; same-day
nonfulfillment costs 100\%. Whole-order failure covers all original
lines, while agreed reduced delivery covers the missing lines.
Unprovided rent is refunded or credited separately, discharging
the corresponding payment liability.

The original basis persists through revisions, and settled obligations
are charged once. Ordinary customer cancellation has no merchant
compensation. A manager can cancel before pickup without waiting
for a risk alert. The evaluated configuration uses contract-level
settlement without an additional global reputation penalty.

\subsection{Terminal Valuation and Runoff}
\label{app:valuation}

Owned asset \(a\) has nominal dealer value
\begin{equation}
 V_a=0.55P_a c_a\max(0,1-w_a)^{0.65},
 \label{eq:dealer}
\end{equation}
subject to monetary rounding, where \(c_a\) is condition and \(w_a\)
cumulative wear. An intact unused asset has a 55\% recovery baseline.

Eligible terminal assets are available, damaged, cleaning, repairing,
or under maintenance. Paid repairs use post-repair condition.
Unpaid cleaning is deducted once from the dealer pool, floored at zero.
Rented, supplier-owned, retired, sold, and unreceived equipment has
zero terminal owned value.

The within-season curves use contract-adjusted value: cash plus equipment
value, less deposit, unearned-revenue, and credit liabilities. This includes
active owned rentals at expected recoverable value, net of expected
cleaning and including expected damage deductions. Prepaid equipment
still in transit is valued at 55\% of its purchase cost. Other owned
equipment uses dealer value less unpaid cleaning. Future reservation rents
are not counted as profit before service. Once contracts and orders are
settled, the curves' terminal endpoints reconcile with the official score.

The fourteen settlement days process existing returns and services,
with no further manager decisions or new pickups. The daily operating
fee stops; holding and outstanding service costs continue.
Refund or compensation shortfalls can draw up to 300 CNY of protective
credit with a 25\% fee. Required-payment insolvency ends the run
with the negative recorded cash shortfall.

%% file: appendices/run_details.tex
\section{Evaluation Protocol and Rule Reference}
\label{app:runs}

\subsection{Shared Study Settings}
\label{app:study-config}

All \StudyRunCount{} model runs share the environment
code, catalog, business instructions, and tool schema. The initial fleet
is fixed across seeds. Table~\ref{tab:environment-config} lists the common
economic settings. Result tables group runs by seed and use the same
model order as Table~\ref{tab:main-results}, with Calendar last.
The warehouse limit is set to 5,000 weighted units, leaving it slack
rather than making storage the principal constraint.

The decision calendar spans April 1--October 17, 2026, followed by automatic
settlement through October 31. Historical Beijing weather is replayed by
month and day. Potential customers are assigned to the reservation channel
with probability 0.7, with booking leads of 1--14 days. Realized channel
shares also depend on customer decisions and service availability.

\paragraph{Weather data in the public release.}
The reported experiments use frozen Visual Crossing weather data. The
public release replaces this input with an openly licensed Beijing series
constructed from NOAA observations and ERA5/ERA5-Land reanalysis, retaining
the same business rules and scoring.

  \par\medskip\begingroup
  \renewenvironment{table}[1][]{%
    \noindent\begin{minipage}{\columnwidth}\centering\captionsetup{type=table}%
  }{\end{minipage}}%
  \input{tables/environment_configuration}  \endgroup\par\medskip

\FloatBarrier

\subsection{Controller, Memory, and Execution}
\label{app:controller}

All runs configure temperature 1.0, an 800,000-character within-day context
budget, 200 responses per day, and 40,000 responses per season. A response
can batch up to five independent read-only queries; state-changing calls
execute individually. Errors and responses without tools consume rounds.
Budget reminders state the remaining rounds, and unfinished days close
automatically at the daily limit.

Business state persists across fresh daily sessions. Named notes can be
searched, read, replaced, or deleted, with a 40,000-character body limit per
document. The agent retrieves notes explicitly; the closing handoff is
supplied automatically the next day. The complete system prompt and tool
inventory appear in Appendix~\ref{app:agent-interface}.

Table~\ref{tab:execution} reports model and tool calls alongside token use.
Model inference costs are separate from the shop's score.

  \begingroup
  \renewenvironment{table*}[1][]{%
    \begin{strip}\begin{minipage}{\textwidth}\centering\captionsetup{type=table}%
  }{\end{minipage}\end{strip}}%
  \input{tables/generated/execution}  \endgroup

\FloatBarrier

\subsection{Aggregation and Inclusion}
\label{app:aggregation}

Each model has one complete run on each of the two seeds. For a run-level
measure \(q\), the reported model mean is
\begin{equation}
 \bar q_m=\frac{q_{m,42}+q_{m,73}}{2}.
 \label{eq:aggregation}
\end{equation}
Rates are computed within each run before averaging. A complete diagnostic
mean requires defined values on both seeds; a missing or inapplicable
observation is not replaced with zero. Counts in the detailed tables retain
their own denominators.

Figure~\ref{fig:mean-value} applies the same averaging at every completed
day to contract-adjusted value minus initial official value. Its final
point equals mean official value creation. All full-season comparisons use
complete, matching-version trajectories. Adding a model leaves the world
generator, metric definitions, and existing model results unchanged.

\subsection{Public-Information Rule Reference}
\label{app:rule-baselines}

We designed Calendar around a fixed assortment and pricing policy, with
daily checks on booking capacity. Its twelve products cover core camping
and picnic needs, with stock targets of two to ten units per product and
first-day tariffs of 3.75--10 CNY per item. These hand-specified settings
remain fixed across days and seeds. Replenishment follows a fixed priority
within cash and capital limits, retaining a 500 CNY reserve net of customer
liabilities. It stops in the last thirty decision days.

The calendar uses current assets, public returns, service-job dates,
inbound estimates, and commitments. A booking is
accepted when estimated rent less ordinary cleaning is positive and it
introduces no projected resource conflict.

Calendar allows one day of ordinary turnaround, evaluates advertised dates
separately, and follows current customer authorization for recovery. It
does not initiate supplier trials or proactive renegotiation. The calendar
is rebuilt daily from public tools under the same 200-round interaction
limit. The policy's internal computation is not token-metered.

Both rule runs finish without insolvency, tool errors, or budget exhaustion;
the maximum daily interaction count is 73. Their environment, data, and
business instructions match the model runs. Their outcomes and channel
revenues appear alongside the models in Appendix~\ref{app:results}.
Extended replay diagnostics cover the LLM runs.

\input{appendices/agent_interface}

%% file: tables/environment_configuration.tex
\begin{table}[tb]
\centering
\small
\begin{tabularx}{\columnwidth}{@{}P{0.40\columnwidth}Y@{}}
\toprule
Component & Evaluated configuration \\
\midrule
Decision / runoff days & \decisiondays{} / \runoffdays{} \\
Initial cash & \initialcash \\
Initial official value & \initialvalue \\
Catalog / initial assets & 46 SKUs / 27 items \\
Initial capital basis & 1,141.83 \\
Additional capital headroom & 1,000.00 \\
Fixed capital ceiling & \capitalceiling \\
Operating fee & 15 per decision day \\
Cleaning capacity & 30 work points per morning \\
Cleaning charge & 0.50 per work point \\
\bottomrule
\end{tabularx}
\caption{Common study configuration. Monetary amounts are simulated CNY;
equipment-capital headroom is distinct from cash.}
\label{tab:environment-config}
\end{table}

%% file: tables/generated/execution.tex
\begin{table*}[tb]
\centering
\small
\setlength{\tabcolsep}{4pt}
\begin{tabular}{@{}lrrrr@{}}
\toprule
Agent & Model calls & Tool calls & Input tokens (M) & Output tokens (M) \\
\midrule
\multicolumn{5}{l}{\textbf{Seed 42}} \\
Sol & 2,524 & 5,835 & 305.26 & 2.58 \\
Luna & 3,279 & 8,850 & 251.81 & 13.63 \\
MiMo & 2,904 & 4,659 & 279.13 & 14.38 \\
Opus & 2,452 & 3,047 & 150.61 & 2.52 \\
Kimi & 2,251 & 3,390 & 138.34 & 4.83 \\
GLM & 2,333 & 3,594 & 118.74 & 14.30 \\
\shortstack[l]{DeepSeek V4.1\\Flash 0910} & 2,964 & 6,897 & 446.61 & 21.22 \\
\multicolumn{5}{l}{\textbf{Seed 73}} \\
Sol & 2,688 & 7,189 & 215.43 & 2.35 \\
Luna & 2,849 & 7,700 & 214.13 & 11.45 \\
MiMo & 2,930 & 4,732 & 289.39 & 14.83 \\
Opus & 2,513 & 3,159 & 162.22 & 2.71 \\
Kimi & 2,241 & 3,374 & 142.36 & 5.15 \\
GLM & 2,317 & 3,605 & 118.95 & 14.92 \\
\shortstack[l]{DeepSeek V4.1\\Flash 0910} & 2,899 & 6,955 & 441.19 & 19.03 \\
\bottomrule
\end{tabular}
\caption{Full-season model and tool calls, with input and output tokens in millions. A model response can request multiple tools, so the call counts differ.}
\label{tab:execution}
\end{table*}

%% file: appendices/agent_interface.tex
\onecolumn
\subsection{Agent Instructions and Tools}
\label{app:agent-interface}

\subsubsection{System Prompt and Daily Inputs}
\label{app:system-prompt}

All evaluated LLM runs use the same system prompt and the same 43 tools.
The prompt below is reproduced verbatim from the saved run artifacts;
only its line wrapping is changed for presentation.

Each day also supplies a morning observation and, when available, the
previous day's handoff. Within a day, the agent receives tool results and
remaining-budget notices. Named notes are retrieved through the memory
tools. Detailed public operating rules are available on demand through
\code{read_business_rules}, under six topics: assets, finance, procurement,
reservations, scoring, and workspace. Thus, the fixed prompt establishes
the task and interaction rules, while tool queries provide detailed rules
and the current business state.

\medskip
\begin{Verbatim}
You manage a seasonal outdoor-equipment rental shop in Beijing. All money is CNY. Keep the shop solvent and maximize its official final value. Use business tools to perform actions; ordinary text and notes do not alter contracts or complete business operations. Detailed public rules are available through read_business_rules; omit topics for its directory or request relevant topics.

Manager decisions, new demand and pickups end on Day 200. Automatic runoff continues through Day 214, when final value is scored; there are no manager actions during runoff. For a solvent shop, final value = cash + final owned-equipment net dealer value - refundable deposits - unearned reservation revenue - refund-protection credit liabilities. An unpaid required payment ends the run in insolvency after any applicable refund protection. See scoring rules for terminal eligibility, deductions and insolvency scoring. Model usage is recorded separately and does not reduce shop cash or the official score.

A newly delivered, unused, intact owned asset has a 55% nominal recovery baseline relative to its recorded acquisition cost. Actual value depends on condition and wear; the full formula and settlement rules are in the scoring handbook. Deposits and unearned prepayments are liabilities, not profit. Cash, durable-capital headroom and warehouse space are separate constraints; capital headroom is not additional cash.

Each day opens after automatic morning processing, including deliveries, returns, service jobs and eligible reservation pickups. Closing resolves pending reservation defaults and pickup risks before walk-ins, charges daily costs and advances the day. Contractual return dates and supplier delivery estimates are not guarantees of physical availability. Customers can continue using rented equipment without approval. Normal returned equipment must pass through cleaning and cannot be rented again on its return day. Current asset status and service terms determine whether it can be used now.

A quotation may create a binding commitment immediately or at close. Customer authorization and successful confirmation do not certify inventory or scheduling feasibility. Every committed line, including a selected optional line, is binding; changes require the applicable customer-authorization process. Merchant cancellation or nonfulfillment incurs refunds/credits and compensation under the disclosed rules. Initial authorization, later modification and actual pickup are different events; current contract terms and settlement receipts record what has occurred. Walk-ins use posted prices, while reservations use their own locked terms. Offers without a posted price do not serve walk-ins. Pricing units and each operation's permissions, deadlines and default consequences are described by its tools and the handbook.

Every business day starts a fresh model session. Business state and named memory documents persist; previous days' reasoning, ordinary assistant text and transcripts are not supplied. Retrieve saved notes through memory tools; they are not automatically loaded. A closing handoff is supplied only on the next day. Older same-day messages may leave the context window.

There are at most 200 model responses per day. Each response consumes a round, including responses without tools or with errors. Budget notices state the remaining rounds, including the next response. At the daily limit, permitted tools from the last response execute before automatic close if the day remains open. You may batch at most 5 independent read-only tools in one response. State changes, memory writes/deletions and close_business_day must each be the only tool request in a response. Invalid batches execute nothing. Tool descriptions identify read-only and state-changing operations. Call close_business_day when today's decisions are complete.
\end{Verbatim}

\clearpage
\subsubsection{Available Tools}
\label{app:tool-inventory}

Table~\ref{tab:agent-tools} lists the complete tool inventory. Tool names and
read/write classifications follow the evaluated interface; the descriptions
summarize their purpose. Read-only queries may be batched as specified in
the prompt. State-changing calls, note updates, and day closing are
submitted individually. Tool descriptions and the public handbook provide
parameter requirements, authorization conditions, deadlines, and default
outcomes.

\begingroup
\small
\setlength{\tabcolsep}{4pt}
\renewcommand{\arraystretch}{1.12}
\input{tables/generated/agent_tools}
\endgroup
\twocolumn

%% file: tables/generated/agent_tools.tex
\begin{longtable}{@{}P{0.37\textwidth}P{0.08\textwidth}P{0.51\textwidth}@{}}
\caption{All 43 tools available to the evaluated agents. R: read-only; W: changes business or memory state; End: closes the business day. Descriptions summarize the evaluated operations.}
\label{tab:agent-tools}\\
\toprule
Tool & Type & Purpose \\
\midrule
\endfirsthead
\multicolumn{3}{@{}l}{\tablename~\thetable{} (continued)}\\
\toprule
Tool & Type & Purpose \\
\midrule
\endhead
\midrule
\multicolumn{3}{r@{}}{\emph{Continued on the next page}}\\
\endfoot
\bottomrule
\endlastfoot
\multicolumn{3}{@{}l}{\textbf{Business records and public information}}\\*
\code{business_status} & R & Inspect finances, obligations, assets, resource limits, and day progress. \\
\code{business_history} & R & Review completed-day business totals and successful actions. \\
\code{finance_history} & R & Review financial totals and daily financial records. \\
\code{service_history} & R & Review rental-service outcomes and product-level performance. \\
\code{counter_log} & R & Read recent individual walk-in interactions and their outcomes. \\
\code{demand_report} & R & Review recorded customer demand and outcomes by selected filters. \\
\code{weather_forecast} & R & Read public weather forecasts and calendar information. \\
\code{read_business_rules} & R & Read the handbook directory or selected public-rule topics. \\
\addlinespace[5pt]
\multicolumn{3}{@{}l}{\textbf{Catalog and procurement}}\\*
\code{search_catalog} & R & Browse supplier categories or search product offers. \\
\code{catalog_report} & R & Filter and inspect a paginated list of supplier offers. \\
\code{inspect_offer} & R & Inspect one product's specifications, costs, delivery estimate, and lifecycle profile. \\
\code{place_purchase_order} & W & Submit a prepaid equipment order or an eligible supplier-trial request. \\
\code{purchase_order_status} & R & Inspect purchase orders, delivery status, and supplier estimates. \\
\addlinespace[5pt]
\multicolumn{3}{@{}l}{\textbf{Equipment, servicing, and asset exits}}\\*
\code{view_rental_assets} & R & Inspect fleet summaries or individual asset condition, service status, and capital basis. \\
\code{view_active_rentals} & R & Read active rentals, assigned equipment, deposits, and expected returns. \\
\code{view_rental_activity} & R & Read equipment event histories with pagination. \\
\code{view_warehouse_capacity} & R & Inspect occupied and available warehouse capacity and product weights. \\
\code{view_liquidation_quotes} & R & Obtain dealer quotes and sale eligibility for current holdings. \\
\code{sell_to_liquidation_dealer} & W & Sell eligible owned assets; capital is released the next business day. \\
\code{repair_or_retire_assets} & W & Order repairs or dispose of damaged equipment. \\
\code{schedule_maintenance} & W & Schedule and prepay maintenance for eligible equipment. \\
\code{return_supplier_trial_assets} & W & Request return of supplier-owned trial equipment. \\
\addlinespace[5pt]
\multicolumn{3}{@{}l}{\textbf{Reservations and customer-authorized changes}}\\*
\code{reservation_inbox} & R & Read current requests, confirmations, customer changes, and pickup risks. \\
\code{view_rental_capacity} & R & Inspect raw asset and contract timelines or current service jobs. \\
\code{view_reservation} & R & Inspect reservation terms, versions, public history, and authorized alternatives. \\
\code{view_reservation_quote_workspace} & R & Inspect request requirements, eligible quotation choices, and fixed-offer plans. \\
\code{quote_reservation_inquiries} & W & Submit a configuration-and-price menu or decline an inquiry; accepted quotes may create commitments. \\
\code{confirm_reservation_plans} & W & Select a customer-authorized plan and permitted pickup date. \\
\code{respond_reservation_service_offers} & W & Accept a complete fixed-term plan and date, or decline the request. \\
\code{respond_reservation_changes} & W & Accept or decline a customer's rescheduling or extension request. \\
\code{resolve_reservation_risks} & W & Request or select authorized recovery options for an exposed pickup risk. \\
\code{cancel_reservation_commitment} & W & Cancel an eligible pre-pickup commitment and settle refunds and compensation. \\
\code{request_reservation_contingency} & W & Propose specific pre-pickup revisions for a next-day customer response. \\
\code{apply_reservation_contingency} & W & Apply an accepted proposal or counteroffer from a valid customer-response report. \\
\addlinespace[5pt]
\multicolumn{3}{@{}l}{\textbf{Walk-in prices and capacity controls}}\\*
\code{set_prices} & W & Set posted walk-in rental prices without changing existing contract terms. \\
\code{set_rental_offer_status} & W & Open or close a product to new walk-in rentals. \\
\code{set_walk_in_capacity_limits} & W & Set or clear product- and date-specific caps on walk-in equipment occupancy. \\
\addlinespace[5pt]
\multicolumn{3}{@{}l}{\textbf{Persistent notes and day closing}}\\*
\code{memory_list} & R & List persistent note names and metadata. \\
\code{memory_search} & R & Search persistent notes using literal keywords. \\
\code{memory_read} & R & Read a named persistent note. \\
\code{memory_write} & W & Create or replace a persistent note. \\
\code{memory_delete} & W & Delete a persistent note. \\
\code{close_business_day} & End & Close the day, execute settlement and defaults, and optionally leave a next-day handoff. \\
\end{longtable}

%% file: appendices/results_details.tex
\section{Detailed Operating Results}
\label{app:results}

\subsection{Outcomes in Each World}
\label{app:world-outcomes}
\label{app:service-outcomes}

Table~\ref{tab:seed-results} gives the individual outcomes underlying the
means in Table~\ref{tab:main-results}. Each seed includes all models and
Calendar, starting from the same initial business value. Financial amounts
are CNY. These results distinguish value created from the volume of
bookings served.

Reservation reliability is
\(100N_{\mathrm{pick}}/(N_{\mathrm{pick}}+N_{\mathrm{fail}})\).
Here \(N_{\mathrm{pick}}\) counts pickups, including agreed reduced
delivery; \(N_{\mathrm{fail}}\) counts manager cancellations, pickup
failures, and merchant-induced customer cancellations. Ordinary customer
cancellations are excluded from this denominator.

  \begingroup
  \renewenvironment{table*}[1][]{%
    \begin{strip}\begin{minipage}{\textwidth}\centering\captionsetup{type=table}%
  }{\end{minipage}\end{strip}}%
  \input{tables/generated/seed_results}  \endgroup

\FloatBarrier

\clearpage
\subsection{Financial Decomposition}
\label{app:financial-decomposition}

Table~\ref{tab:financial-decomposition} reconciles value creation with
revenue and expenditure. Net equipment outlay is purchases less dealer-sale
proceeds and final owned-equipment value.
Operating expenses include cleaning, repair, maintenance, holding,
reservation compensation, supplier-trial charges, and the fixed daily fee.
The three income columns, less net equipment outlay, operating expenses,
and the common initial equipment value of 397.06 CNY, give the value gains
in Table~\ref{tab:seed-results}. Completed runs have settled their deposit,
unearned-revenue, and debt liabilities.

For example, Sol seed 73 has net equipment outlay of
\(2070.45-521.29-525.36=1023.80\) CNY. This accounting uses actual expenses
and recovered equipment value; the estimated use consumption in capital
productivity is not an additional expense here.

  \begingroup
  \renewenvironment{table*}[1][]{%
    \begin{strip}\begin{minipage}{\textwidth}\centering\captionsetup{type=table}%
  }{\end{minipage}\end{strip}}%
  \input{tables/generated/financial_decomposition}  \endgroup

\FloatBarrier

\clearpage
\subsection{Value and Financial Paths}
\label{app:financial-paths}

Figure~\ref{fig:value-paths} retains the individual paths behind
Figure~\ref{fig:mean-value}. Each panel compares agents within one seed,
using common axes. Within-season values account for outstanding contracts;
the valuation is defined in Appendix~\ref{app:valuation}.
Terminal endpoints equal the official value gains in
Table~\ref{tab:seed-results}.

\begin{appendixfigure}
\centering
\includegraphics[width=\textwidth]{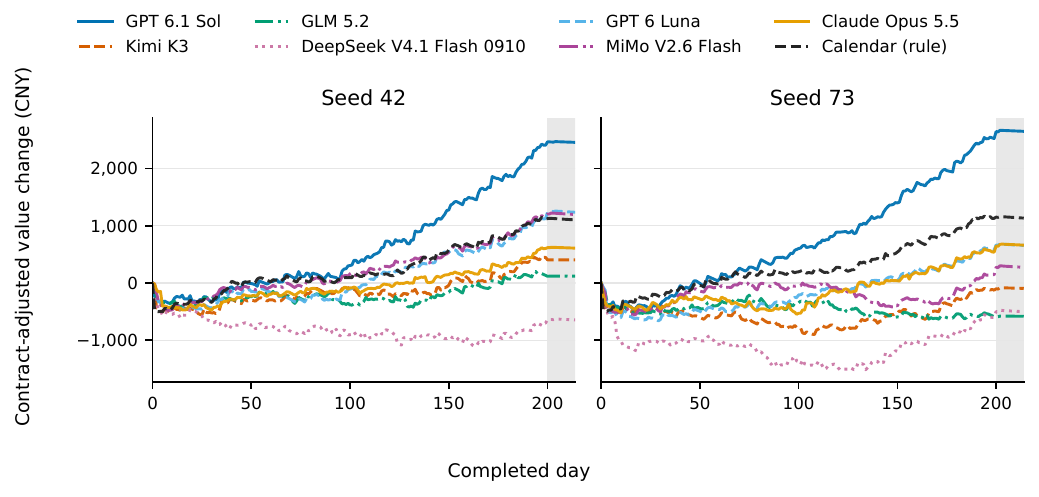}
\caption{Individual value paths, grouped by seed. Curves show
contract-adjusted economic value minus initial business value; endpoints
equal official net value creation. The black line is Calendar. Shading
marks automatic settlement after manager decisions end.}
\label{fig:value-paths}
\end{appendixfigure}
\FloatBarrier

\clearpage
Figures~\ref{fig:purchase-paths} and~\ref{fig:compensation-paths} show
equipment purchases and compensation over the same seasons. DeepSeek
purchases more equipment than Sol in both worlds and pays more
compensation. Purchases can subsequently produce rents, sale proceeds,
and terminal value; the financial decomposition above includes those
recoveries.

\begin{appendixfigure}
\centering
\includegraphics[width=\textwidth]{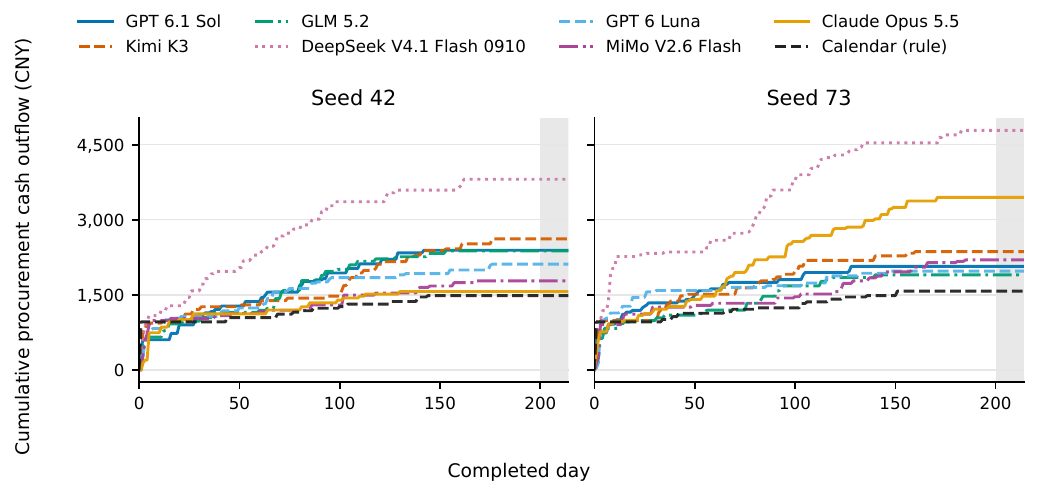}
\caption{Cumulative equipment purchases by seed. Capital released from
asset exits can be reinvested, so cumulative purchases may exceed the
fixed simultaneous capital ceiling. Shading marks automatic settlement.}
\label{fig:purchase-paths}
\end{appendixfigure}

\begin{appendixfigure}
\centering
\includegraphics[width=\textwidth]{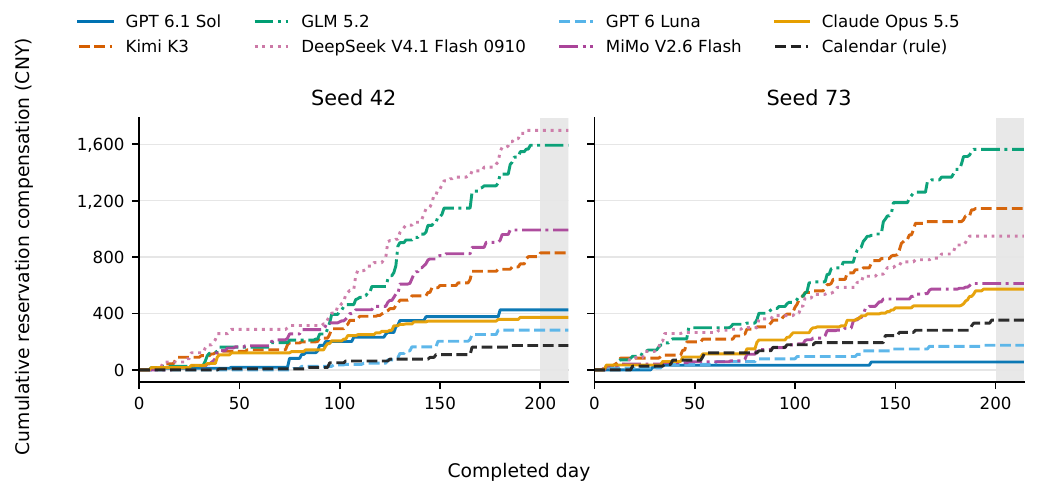}
\caption{Cumulative compensation for reservation changes and failures by
seed. Refunds of unearned payments are recorded separately. Shading marks
automatic settlement.}
\label{fig:compensation-paths}
\end{appendixfigure}
\FloatBarrier

%% file: tables/generated/seed_results.tex
\begin{table*}[tb]
\centering
\small
\setlength{\tabcolsep}{4pt}
\begin{tabular}{@{}lrrrr@{}}
\toprule
Agent & $\Delta V$ & Bookings & Reliab. (\%) & Compens. \\
\midrule
\multicolumn{5}{l}{\textbf{Seed 42}} \\
Sol & +2,460.38 & 263 & 92.93 & 426.33 \\
Luna & +1,237.24 & 188 & 92.16 & 282.12 \\
MiMo & +1,198.11 & 266 & 81.85 & 992.47 \\
Opus & +610.03 & 289 & 91.46 & 372.64 \\
Kimi & +403.26 & 221 & 81.85 & 829.57 \\
GLM & +119.63 & 244 & 72.19 & 1,592.91 \\
\shortstack[l]{DeepSeek V4.1\\Flash 0910} & -644.71 & 296 & 74.56 & 1,699.12 \\
Calendar & +1,103.57 & 168 & 93.85 & 173.46 \\
\multicolumn{5}{l}{\textbf{Seed 73}} \\
Sol & +2,653.18 & 244 & 98.79 & 56.11 \\
Luna & +657.63 & 225 & 95.34 & 175.23 \\
MiMo & +274.47 & 260 & 87.25 & 612.81 \\
Opus & +662.19 & 304 & 89.94 & 572.77 \\
Kimi & -94.17 & 256 & 79.50 & 1,144.43 \\
GLM & -581.95 & 243 & 72.97 & 1,563.61 \\
\shortstack[l]{DeepSeek V4.1\\Flash 0910} & -501.88 & 293 & 83.71 & 949.03 \\
Calendar & +1,135.45 & 145 & 89.51 & 353.24 \\
\bottomrule
\end{tabular}
\caption{Full-season outcomes within each world. Monetary values are CNY; $\Delta V$ is the change from the initial business value of 3,397.06. Bookings are reservation pickups, including agreed reduced delivery; reliability is pickups divided by pickups plus merchant-caused failures.}
\label{tab:seed-results}
\end{table*}

%% file: tables/generated/financial_decomposition.tex
\begin{table*}[tb]
\centering
\small
\setlength{\tabcolsep}{4pt}
\begin{tabular}{@{}lrrrrr@{}}
\toprule
Agent & \shortstack{Reservation\\rent} & \shortstack{Walk-in\\rent} & \shortstack{Damage\\deductions} & \shortstack{Net equipment\\outlay} & \shortstack{Operating\\expenses} \\
\midrule
\multicolumn{6}{l}{\textbf{Seed 42}} \\
Sol & 7,344.36 & 2,609.75 & 218.98 & 1,223.21 & 6,092.44 \\
Luna & 5,326.23 & 2,510.27 & 193.49 & 1,024.41 & 5,371.28 \\
MiMo & 6,058.99 & 2,949.25 & 283.20 & 1,002.19 & 6,694.08 \\
Opus & 5,677.90 & 1,818.25 & 318.56 & 834.22 & 5,973.40 \\
Kimi & 5,682.52 & 2,340.00 & 237.78 & 1,283.25 & 6,176.73 \\
GLM & 5,723.16 & 2,949.50 & 305.85 & 1,256.71 & 7,205.11 \\
\shortstack[l]{DeepSeek V4.1\\Flash 0910} & 6,768.39 & 2,228.60 & 252.99 & 1,991.40 & 7,506.23 \\
Calendar & 3,313.74 & 4,141.08 & 153.01 & 732.45 & 5,374.75 \\
\multicolumn{6}{l}{\textbf{Seed 73}} \\
Sol & 6,513.78 & 2,964.00 & 209.65 & 1,023.80 & 5,613.39 \\
Luna & 4,924.24 & 2,449.76 & 176.80 & 951.70 & 5,544.41 \\
MiMo & 5,316.81 & 2,461.50 & 288.42 & 1,254.49 & 6,140.71 \\
Opus & 6,579.99 & 2,506.55 & 234.96 & 1,800.24 & 6,462.01 \\
Kimi & 6,083.16 & 1,972.75 & 217.56 & 1,278.31 & 6,692.27 \\
GLM & 5,012.81 & 2,814.00 & 208.36 & 1,085.70 & 7,134.36 \\
\shortstack[l]{DeepSeek V4.1\\Flash 0910} & 6,336.27 & 2,520.25 & 169.01 & 2,388.18 & 6,742.17 \\
Calendar & 3,059.61 & 4,741.76 & 289.12 & 814.20 & 5,743.78 \\
\bottomrule
\end{tabular}
\caption{Full-season income and expenditure in CNY. The three income columns, less net equipment outlay, operating expenses, and the common initial equipment value of 397.06, yield $\Delta V$ in Table~\ref{tab:seed-results}. Estimated usage wear is not an additional expense here.}
\label{tab:financial-decomposition}
\end{table*}

%% file: appendices/diagnostics.tex
\section{Diagnostic Methods and Detailed Results}
\label{app:diagnostics}

The four families follow Figure~\ref{fig:behavioral-overview}A--D, with
definitions and results together. Results are grouped by seed.
These diagnostics cover the LLM runs; Calendar's operating results appear
in Appendix~\ref{app:results}.

\subsection{Observation Windows and Readiness Projection}
\label{app:readiness}

Tables cover the complete season and retain numerators and denominators.
Model means average two run-level measurements
(Appendix~\ref{app:aggregation}).

The pending-booking and action-induced-gap diagnostics project supply
from public asset states and schedules. Ready units are available today.
FIFO cleaning completes after cumulative remaining work divided by daily
capacity, rounded up to at least one more morning. Repair and maintenance
use their scheduled ready dates.

Rented equipment uses the latest public estimated return, or its approved
or promised return, plus the turnaround allowance \(g\). Overdue assets
without a future return estimate remain outside projected supply. Inbound
owned orders use a public ETA later than today. Trial supply ends at its
deadline and is excluded once return is pending or no rental starts remain.
The booking calendar includes confirmed and at-risk commitments, staged
products, and authorized omissions.

A booking occupies product capacity over \([p,r+g)\), with pickup \(p\)
and scheduled return \(r\). The standard allowance is \(g=1\). Checks at
day \(t\) compare product quantities on dates from \(t\) onward. A booking
has a gap if total demand exceeds projected supply in any product--date
cell it occupies. All bookings using that cell are marked, rather than
allocating the shortfall by priority. Future
cleaning congestion, damage, late returns, and service-life exhaustion can
change actual availability; later information can also restore supply
excluded by an earlier projection.

\subsection{Pending-Booking Gaps}
\label{app:pending-gaps}

After each day's walk-ins, the evaluator checks pending bookings with
future pickup dates. For observed booking--days \(\mathcal H\), the rate is
\begin{equation}
 100\frac{\sum_{(c,t)\in\mathcal H}
 \indicator[\text{\(c\) has a projected gap at \(t\)}]}{|\mathcal H|}.
\end{equation}
Each booking retains its identity across revisions and may contribute on
multiple days. A gap can end through improved projected supply, pickup, or
cancellation. The measure therefore describes support for retained
commitments, alongside the delivery outcomes in
Appendix~\ref{app:service-outcomes}.

Table~\ref{tab:gap-lead-counts} gives the overall and near-pickup results
behind Figure~\ref{fig:behavioral-overview}A. Near-pickup observations have
one or two days until pickup; this is not a fixed cohort followed through
recovery.

%
  \par\medskip\begingroup
  \renewenvironment{table}[1][]{%
    \noindent\begin{minipage}{\columnwidth}\centering\captionsetup{type=table}%
  }{\end{minipage}}%
  \input{tables/generated/gap_lead_counts}  \endgroup\par\medskip

\FloatBarrier

\clearpage
\subsection{Action-Induced New Gaps}
\label{app:action-gaps}

For each supported successful action, replay compares projected coverage
immediately before and after execution. An action is relevant when it
changes a product used by a pre-existing pending booking. It introduces a
new gap when a booking present on both sides changes from covered to short.
The rate counts gap-introducing calls over relevant calls, separately for
each action type. One call counts once even if several bookings become
short; removing a booking is not counted as creating a gap for that booking.

Walk-in rentals are checked separately. Actual rentals are grouped by
customer trip and added sequentially to the pre-walk-in calendar. A batch
counts when it introduces a shortage for a previously covered booking.
Its denominator is all actual walk-in rental batches, rather than manager
tool calls. The same one-day turnaround allowance is used.

Table~\ref{tab:action-gap-main} gathers the four series in
Figure~\ref{fig:behavioral-overview}B, keeping their different exposure
denominators separate.

  \begingroup
  \renewenvironment{table*}[1][]{%
    \begin{strip}\begin{minipage}{\textwidth}\centering\captionsetup{type=table}%
  }{\end{minipage}\end{strip}}%
  \input{tables/generated/action_gap_main}  \endgroup

\FloatBarrier

\clearpage
\subsection{Capital Productivity}
\label{app:capital-metric}

Let \(\Pi^{\mathrm{cash}}\) be rental revenue and damage deductions less
cleaning, repair, maintenance, holding, reservation compensation, and
supplier-trial fees over the season. This service contribution excludes
fixed daily operating fees and equipment purchase/sale cash flows.

For an owned asset \(a\), a rental of \(d\) scheduled days has estimated
linear use consumption
\begin{equation}
 u_a(d)=P_a\left(\frac{d}{L_a^d}+\frac{1}{L_a^c}\right),
 \label{eq:use-consumption}
\end{equation}
where \(P_a\) is acquisition cost and \(L_a^d,L_a^c\) are the public day
and cycle limits. The cycle term is omitted when absent. Each additional
possession day adds \(P_a/L_a^d\). Recorded daily consumption is used where
available and otherwise reconstructed with the same rounding. Denote the
seasonal total by \(U^{\mathrm{use}}\).

Using the capital quantities in Appendix~\ref{app:capital},
\begin{equation}
 Q^{\mathrm{cap}}=
 100\frac{\Pi^{\mathrm{cash}}-U^{\mathrm{use}}}
 {\sum_{t=1}^{T}(B_t+I_t)}.
 \label{eq:capital-productivity}
\end{equation}
The denominator accumulates day-end occupied capital, including inbound
orders and basis awaiting release. The unit is CNY per 100 capital-CNY-days.
Trial equipment contributes service income and fees without owned basis
or an owned-wear charge.

Table~\ref{tab:capital-details} provides the components and the individual
values behind Figure~\ref{fig:behavioral-overview}C. This measure describes
returns per occupied capital-time; Appendix~\ref{app:financial-decomposition}
separately reconciles total business value using actual costs and dealer
values.

  \begingroup
  \renewenvironment{table*}[1][]{%
    \begin{strip}\begin{minipage}{\textwidth}\centering\captionsetup{type=table}%
  }{\end{minipage}\end{strip}}%
  \input{tables/generated/capital_details}  \endgroup

\FloatBarrier

\clearpage
\subsection{Unrealized Gains from Capacity Relaxation}
\label{app:relaxation}

The test asks whether a small relaxation of a walk-in limit improves
estimated service contribution under a conservative supply check. It
samples days \(5,10,\ldots,200\): forty dates fixed independently of model
results.

A candidate raises one product's existing finite occupancy limits by one
over configured dates in \([t,t+14]\), clipped to the season. In a copy of
the pre-walk-in state, the evaluator applies the capacity tool and reruns
that day's walk-ins with the same stock, prices, commitments, and customers.
Only products with finite limits contribute candidates; the set therefore
depends on the agent's fleet and prior decisions.

Let \(J_t(a)\) be rent agreed in the resulting rentals less normal cleaning
and owned-use consumption or trial fees. The estimate uses information at
the branch, excluding future damage and late-return outcomes. The unchanged
branch must reproduce the observed transactions.

Safety allowances \(g\in\{1,3,7\}\) are tested separately. Walk-ins and
bookings occupy \([p,r+g)\), reserving \(g\) turnaround days starting on
the scheduled return day \(r\). Wherever a
candidate increases occupancy, walk-in use and existing booking demand
must fit the quantity available at branch start. This test reserves
capacity without anticipating new returns or purchases.

For candidates \(\mathcal A_t^g\) that pass, estimated improvement is
\begin{equation}
 \begin{aligned}
 \Delta_t^g=\max\bigl(&\{0\}\cup\\
 &\{J_t(a)-J_t(0):a\in\mathcal A_t^g\}\bigr).
 \end{aligned}
 \label{eq:local-relaxation}
\end{equation}
The mean includes all forty sampled dates, including those with no
improving candidate. An improving date has a gain of at least 0.01 CNY
and counts once even when several products qualify.

Table~\ref{tab:relaxation} reports the \(g=3\) results used in
Figure~\ref{fig:behavioral-overview}D alongside the \(g=1,7\) checks.
A larger value indicates more estimated gains left in the recorded
configuration; a smaller value alone does not establish better allocation.
Each branch changes one product's limits over the specified date range
but reruns only the sampled day's walk-ins.
Subsequent policy decisions are not rerun, so these estimates cannot be
added into recoverable season profit. Appendix~\ref{app:capacity-cases}
shows the rental intervals behind representative positive tests.

  \begingroup
  \renewenvironment{table*}[1][]{%
    \begin{strip}\begin{minipage}{\textwidth}\centering\captionsetup{type=table}%
  }{\end{minipage}\end{strip}}%
  \input{tables/generated/relaxation}  \endgroup

\FloatBarrier

\clearpage
\subsection{Sensitivity to Turnaround in the Nominal Calendar}
\label{app:calendar-sensitivity}

This supplementary check asks how tightly new bookings are spaced under
different turnaround assumptions. It uses the closing book on the day of
confirmation, rather than the public-readiness projection above.
Supply consists of initial owned units and accepted purchases, less exits.
Owned orders count from acceptance; delivered trial units count until
return or their deadline. Physical delivery, cleaning congestion, walk-ins,
damage, and later service-life exhaustion are abstracted out.

A new booking is covered when quantity fits at every product--date cell it
uses. Table~\ref{tab:calendar-sensitivity} reports coverage for \(g=0,1,2\),
where \(g=0\) permits return-day reuse. The change from \(g=0\) to \(g=1\)
supports the comparison in Section~\ref{sec:finding-continuity}.
Bookings terminated on the confirmation day are excluded. Procurement
and revisions completed that day affect the result.

  \begingroup
  \renewenvironment{table*}[1][]{%
    \begin{strip}\begin{minipage}{\textwidth}\centering\captionsetup{type=table}%
  }{\end{minipage}\end{strip}}%
  \input{tables/generated/calendar_sensitivity}  \endgroup

\FloatBarrier

\clearpage
\subsection{Feasibility and Resource Returns Within Each World}
\label{app:diagnostic-relationship}

Figure~\ref{fig:behavioral-by-seed} relates pending-booking gaps to capital
productivity within each seed. It complements the mean profile in
Figure~\ref{fig:behavioral-overview}: comparatively reliable schedules can
still produce different returns from occupied capital. The panels retain
world-specific variation using identical axes.

\input{figures/behavioral_by_seed}
\FloatBarrier

\subsection{Replay Integrity}
\label{app:replay}

Replay uses the recorded environment version and runtime data and checks
executed actions against their recorded receipts. Unchanged local branches
must reproduce the relevant transactions and balances before alternatives
are evaluated. A mismatch or incomplete branch stops the calculation.
The detailed tables are generated from these completed reports; the
presentation layer does not run new agent decisions.

%% file: tables/generated/gap_lead_counts.tex
\begin{table}[tb]
\centering
\small
\setlength{\tabcolsep}{3pt}
\begin{tabular}{@{}lrr@{}}
\toprule
Agent & \shortstack{All pending\\bookings} & \shortstack{1--2 days\\before pickup} \\
\midrule
\multicolumn{3}{l}{\textbf{Seed 42}} \\
Sol & 8.76 (198/2260) & 13.70 (74/540) \\
Luna & 9.17 (141/1537) & 12.18 (47/386) \\
MiMo & 26.52 (590/2225) & 34.59 (211/610) \\
Opus & 5.96 (153/2565) & 8.83 (53/600) \\
Kimi & 22.56 (446/1977) & 30.24 (150/496) \\
GLM & 34.10 (787/2308) & 45.63 (277/607) \\
\shortstack[l]{DeepSeek V4.1\\Flash 0910} & 40.51 (1121/2767) & 52.19 (381/730) \\
\multicolumn{3}{l}{\textbf{Seed 73}} \\
Sol & 3.19 (59/1848) & 2.12 (10/471) \\
Luna & 8.65 (147/1699) & 6.87 (31/451) \\
MiMo & 16.40 (342/2085) & 18.92 (105/555) \\
Opus & 8.95 (240/2683) & 10.73 (68/634) \\
Kimi & 22.40 (515/2299) & 30.19 (176/583) \\
GLM & 27.68 (599/2164) & 40.75 (240/589) \\
\shortstack[l]{DeepSeek V4.1\\Flash 0910} & 26.59 (660/2482) & 31.55 (207/656) \\
\bottomrule
\end{tabular}
\caption{Projected gaps: percentage (gapped/observed booking--days). A booking can contribute on multiple days. The near-pickup subset is determined at each observation, not a fixed recovery cohort.}
\label{tab:gap-lead-counts}
\end{table}

%% file: tables/generated/action_gap_main.tex
\begin{table*}[tb]
\centering
\small
\setlength{\tabcolsep}{4pt}
\begin{tabular}{@{}lrrrr@{}}
\toprule
Agent & Confirm & Quote & Fixed offer & Walk-in \\
\midrule
\multicolumn{5}{l}{\textbf{Seed 42}} \\
Sol & 3.13 (3/96) & 3.33 (2/60) & 6.98 (6/86) & 0.50 (1/201) \\
Luna & 7.50 (3/40) & 8.16 (4/49) & 2.63 (2/76) & 2.23 (4/179) \\
MiMo & 15.22 (14/92) & 11.29 (7/62) & 9.09 (8/88) & 4.13 (10/242) \\
Opus & 1.55 (2/129) & 2.74 (2/73) & 4.48 (3/67) & 1.76 (3/170) \\
Kimi & 23.68 (18/76) & 14.58 (7/48) & 10.98 (9/82) & 3.28 (6/183) \\
GLM & 16.67 (12/72) & 14.29 (11/77) & 21.28 (20/94) & 6.28 (15/239) \\
\shortstack[l]{DeepSeek V4.1\\Flash 0910} & 26.79 (30/112) & 21.13 (15/71) & 26.32 (25/95) & 5.10 (10/196) \\
\multicolumn{5}{l}{\textbf{Seed 73}} \\
Sol & 0.00 (0/87) & 0.00 (0/44) & 1.22 (1/82) & 0.00 (0/228) \\
Luna & 1.89 (1/53) & 2.82 (2/71) & 1.41 (1/71) & 2.83 (6/212) \\
MiMo & 12.50 (12/96) & 7.41 (4/54) & 6.67 (5/75) & 2.19 (4/183) \\
Opus & 2.83 (3/106) & 2.50 (2/80) & 9.21 (7/76) & 0.93 (2/215) \\
Kimi & 17.24 (15/87) & 13.33 (8/60) & 20.48 (17/83) & 4.93 (10/203) \\
GLM & 16.49 (16/97) & 20.55 (15/73) & 14.81 (12/81) & 8.37 (20/239) \\
\shortstack[l]{DeepSeek V4.1\\Flash 0910} & 12.39 (14/113) & 8.20 (5/61) & 16.05 (13/81) & 4.24 (10/236) \\
\bottomrule
\end{tabular}
\caption{Individual results behind Figure~\ref{fig:behavioral-overview}B: percentage (gap-introducing/exposed units). Confirm, Quote, and Fixed offer use relevant successful calls; Walk-in uses actual customer rental batches. Each unit counts once even if several bookings acquire gaps. Types are not pooled. Definitions are in Appendix~\ref{app:action-gaps}.}
\label{tab:action-gap-main}
\end{table*}

%% file: tables/generated/capital_details.tex
\begin{table*}[tb]
\centering
\small
\setlength{\tabcolsep}{4pt}
\begin{tabular}{@{}lrrr@{}}
\toprule
Agent & \shortstack{Net service contribution\\(CNY)} & \shortstack{Occupied capital\\(CNY-days)} & Productivity \\
\midrule
\multicolumn{4}{l}{\textbf{Seed 42}} \\
Sol & 5,917.79 & 430,562.99 & 1.374 \\
Luna & 4,687.77 & 448,281.77 & 1.046 \\
MiMo & 4,430.20 & 443,423.04 & 0.999 \\
Opus & 3,820.33 & 421,113.66 & 0.907 \\
Kimi & 4,048.36 & 419,729.18 & 0.965 \\
GLM & 3,622.30 & 405,994.63 & 0.892 \\
\shortstack[l]{DeepSeek V4.1\\Flash 0910} & 3,289.89 & 441,049.65 & 0.746 \\
\multicolumn{4}{l}{\textbf{Seed 73}} \\
Sol & 6,009.77 & 431,241.55 & 1.394 \\
Luna & 4,028.44 & 450,966.91 & 0.893 \\
MiMo & 3,634.71 & 445,618.46 & 0.816 \\
Opus & 4,494.90 & 440,242.94 & 1.021 \\
Kimi & 3,349.56 & 429,765.80 & 0.779 \\
GLM & 2,680.66 & 406,125.62 & 0.660 \\
\shortstack[l]{DeepSeek V4.1\\Flash 0910} & 3,881.40 & 447,145.31 & 0.868 \\
\bottomrule
\end{tabular}
\caption{Capital productivity: net service contribution per 100 capital-CNY-days. Contribution deducts estimated use consumption but excludes fixed operating fees and equipment purchase/sale cash flows. Capital-days include basis awaiting next-day release.}
\label{tab:capital-details}
\end{table*}

%% file: tables/generated/relaxation.tex
\begin{table*}[tb]
\centering
\small
\setlength{\tabcolsep}{4pt}
\begin{tabular}{@{}lrrrrrr@{}}
\toprule
Agent & \shortstack{$g=1$\\Mean gain} & Days & \shortstack{$g=3$\\Mean gain} & Days & \shortstack{$g=7$\\Mean gain} & Days \\
\midrule
\multicolumn{7}{l}{\textbf{Seed 42}} \\
Sol & 1.210 & 9 & 0.654 & 5 & 0.064 & 1 \\
Luna & 2.190 & 13 & 1.556 & 10 & 0.667 & 4 \\
MiMo & 1.217 & 12 & 0.500 & 6 & 0.453 & 5 \\
Opus & 1.007 & 11 & 0.609 & 7 & 0.227 & 3 \\
Kimi & 1.727 & 14 & 0.929 & 9 & 0.623 & 5 \\
GLM & 1.195 & 11 & 0.792 & 9 & 0.302 & 4 \\
\shortstack[l]{DeepSeek V4.1\\Flash 0910} & 0.827 & 10 & 0.428 & 6 & 0.184 & 3 \\
\multicolumn{7}{l}{\textbf{Seed 73}} \\
Sol & 0.941 & 7 & 0.509 & 5 & 0.052 & 1 \\
Luna & 1.894 & 17 & 1.776 & 15 & 0.775 & 8 \\
MiMo & 0.350 & 2 & 0.000 & 0 & 0.000 & 0 \\
Opus & 0.654 & 6 & 0.242 & 2 & 0.186 & 1 \\
Kimi & 1.660 & 17 & 0.704 & 9 & 0.190 & 2 \\
GLM & 1.106 & 9 & 0.692 & 6 & 0.316 & 2 \\
\shortstack[l]{DeepSeek V4.1\\Flash 0910} & 0.600 & 8 & 0.279 & 4 & 0.000 & 0 \\
\bottomrule
\end{tabular}
\caption{Capacity-relaxation tests on 40 sampled days per run. For each turnaround allowance, Mean gain is estimated CNY improvement averaged over all sampled days, including zero and no-candidate days; Days counts those with an improvement of at least 0.01 CNY. The main figure uses $g=3$.}
\label{tab:relaxation}
\end{table*}

%% file: tables/generated/calendar_sensitivity.tex
\begin{table*}[tb]
\centering
\small
\setlength{\tabcolsep}{4pt}
\begin{tabular}{@{}lrrrr@{}}
\toprule
Agent & Bookings checked & $g=0$ (\%) & $g=1$ (\%) & $g=2$ (\%) \\
\midrule
\multicolumn{5}{l}{\textbf{Seed 42}} \\
Sol & 302 & 98.01 & 95.70 & 78.15 \\
Luna & 215 & 96.74 & 93.02 & 78.60 \\
MiMo & 339 & 94.99 & 84.96 & 67.26 \\
Opus & 337 & 99.41 & 98.81 & 72.11 \\
Kimi & 285 & 95.09 & 82.46 & 67.02 \\
GLM & 350 & 92.57 & 76.57 & 62.86 \\
\shortstack[l]{DeepSeek V4.1\\Flash 0910} & 415 & 89.64 & 68.19 & 53.73 \\
\multicolumn{5}{l}{\textbf{Seed 73}} \\
Sol & 257 & 99.22 & 99.22 & 96.11 \\
Luna & 240 & 97.08 & 96.67 & 85.42 \\
MiMo & 309 & 95.15 & 90.29 & 69.26 \\
Opus & 354 & 96.33 & 94.63 & 74.58 \\
Kimi & 333 & 94.89 & 84.68 & 61.86 \\
GLM & 342 & 90.64 & 80.41 & 66.67 \\
\shortstack[l]{DeepSeek V4.1\\Flash 0910} & 361 & 91.97 & 80.61 & 65.37 \\
\bottomrule
\end{tabular}
\caption{New-booking calendar coverage at day end under three turnaround allowances. The checked set is identical across allowances within each run; confirmations terminated that day are excluded.}
\label{tab:calendar-sensitivity}
\end{table*}

%% file: figures/behavioral_by_seed.tex
\begin{appendixfigure}
\centering
\includegraphics[width=\textwidth]{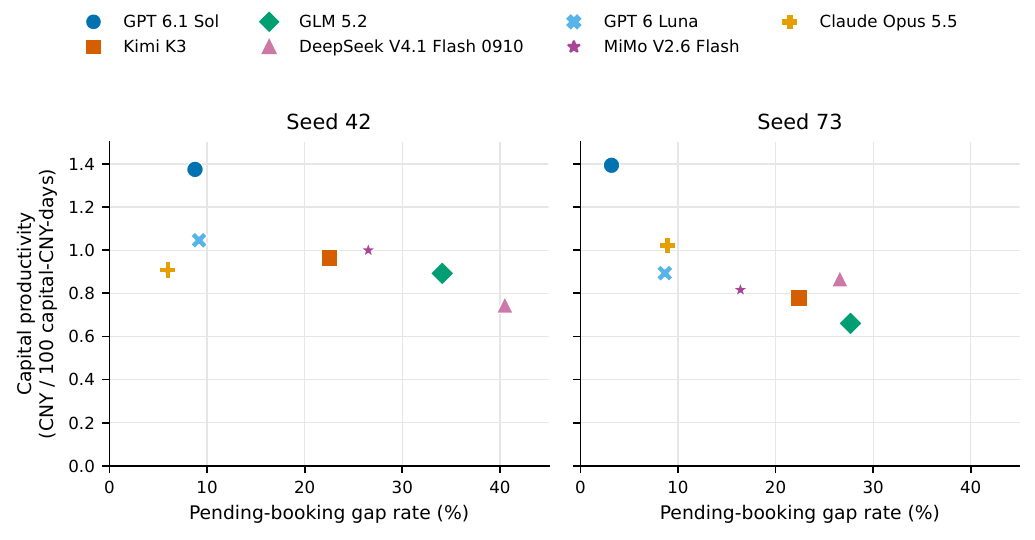}
\caption{Pending-booking gaps and capital productivity within each world.
Each point is one complete LLM run; both panels use identical scales.
These are the individual values underlying panels A and C of
Figure~\ref{fig:behavioral-overview}.}
\label{fig:behavioral-by-seed}
\end{appendixfigure}

%% file: appendices/case_evidence.tex
\section{Evidence for the Trajectory Analysis}
\label{app:cases}

The subsections follow the two themes and five topics in
Sections~\ref{sec:finding-continuity}--\ref{sec:finding-value}.
Notes, handoffs, and visible responses record
the agent's assessment; receipts and asset histories establish actions
and outcomes. Dates are simulation days and amounts are CNY.
Local tests follow Appendix~\ref{app:diagnostics}.

\subsection{Intertemporal Feasibility Maintenance}
\label{app:continuity-cases}

\subsubsection{Applying Operating Rules Consistently}
\label{app:rule-continuity}

In seed 42, DeepSeek had recorded and applied the cleaning rule by day 39,
including cancelling bookings 054 and 064. On days 144--145, however, it
again planned to use sleeping pads on their return day for bookings 290 and 296.
Two sleeping pads returned on time but became ready the following day; the
cancellations cost 14.80 in total.

The later chair review makes the change in its working rule explicit.
Booking 383 was already confirmed on day 177. On day 178, DeepSeek first
counts one ready chair against three required. Seeing the pickup confirmed
for the returning chairs' return date, it then treats confirmation as
evidence that same-day reuse is permitted. It records the new interpretation
and carries it into the day-179 note. The chairs return normally on day 180
but finish cleaning on 181, missing the pickup. Cancellation costs 16,
with a separate refund of the 4 prepayment. This sequence concerns
reassessment of an existing booking: the agent had previously applied the
rule correctly, then changed it while interpreting its own commitment.

A separate occurrence appears in seed 73. DeepSeek revises a four-table
estimate for day 19 to two after recognizing that the returning tables
remain unavailable, followed by cancellation of bookings 010 and 024.
Together, the records show recurrence across both runs.

\subsubsection{Coordinating New Plans with Existing Commitments}
\label{app:joint-planning}
\label{app:operating-process}

\paragraph{Preparing supply across financial and physical delays.}
In Sol seed 73, a revised cookware schedule leaves a returning unit
unavailable for another pickup until cleaning is complete. The shop needs
a fourth unit. Cash is sufficient, but equipment capital is committed.
Sol sells a worn mat on day 23, buys the 27.90 cookware unit after capital
is released on 24, and receives it on 30 rather than the estimated 28.
The new unit supplies booking 026 on day 33. The sequence connects a
future shortage to an asset exit, next-day funding, procurement, and
a delivery that still arrives before the obligation.

Sale--release--purchase sequences recur in Sol seed 42 on days 40--41,
52--53, 60--61, and 128--129. The capital choices in
Appendix~\ref{app:capital-comparisons} provide additional seed-73 sequences.
This execution pattern also occurs in Kimi seed 42: after a rejected date
proposal, it sells a chair on day 152 for 14.30 and buys a 32.61 mat after
capital release on 153. Arrival on 158 supplies two day-165 bookings
with rents of 39 and 13 and no compensation.

\paragraph{Applying the same source of readiness information.}
Opus seed 42 receives a day-128 counteroffer on day 124 for booking 149.
A cleaning-queue check shows that the table and chairs will be ready for
the original day-125 pickup, which it retains and fulfills.
On day 129, another queue query shows the sleeping bags and pads for
booking 193 ready on day 131, after its day-130 pickup. Opus continues
preparing for a chair shortage on day 133 without changing booking 193's
supply plan. The booking is cancelled on 130, with 10 compensation and
a separate 2.50 prepayment refund. Both reviews obtain relevant queue
information; they differ in which obligations that information governs.

\paragraph{Accepting an overlapping commitment.}
In Sol seed 42, booking 093 is accepted on day 78 for days 90--92 and
booking 097 on day 82 for days 89--92. Both require the shop's sole
long-runtime lantern. Before accepting 097, Sol queries the prior booking.
Its visible responses then describe the new offer as coverable and
supportable. It recognizes the double booking on day 88.
The overlap exists during rental, independently of cleaning assumptions.

A local check reproduces the preceding actions: accepting 097 introduces
a shortage for both bookings, while declining it leaves 093 covered on
that product. Renegotiation for 093 fails; 097 takes the lantern on day 89,
and 093 is cancelled on 90 with 40.50 compensation. The pre-acceptance
record indicates a feasibility misjudgment rather than a deliberate
economic choice to exit the earlier obligation.

\paragraph{Maintaining protection after acceptance.}
Sol seed 42 accepts booking 086 on day 72 for a tent on days 83--86.
That tent is rented to a walk-in on day 81, returns normally on 84,
and finishes cleaning on 85. Booking 086 is cancelled on 83 with 12.53
compensation. The walk-in also earns revenue; the sequence identifies an
unprotected obligation rather than a net-profit estimate for refusing
that customer.

Other records show protection being updated with supply. In seed 73,
Sol follows day-23 and day-24 acceptances with future occupancy limits,
while reopening a one-day window for a chair and mat. On day 183, after
customers refuse date changes, it counts six ready comfort chairs and four
finishing cleaning before seven are required the next morning. It opens
two chairs to current one-day walk-ins and sets the next day's limit to
zero. All five day-184 bookings are delivered. No extra chair rental occurs
on 183, so this episode documents executed capacity adjustment rather
than realized additional income.

\subsection{Intertemporal Resource Valuation and Reallocation}

\subsubsection{Choosing Which Commitments to Make and Retain}
\label{app:matched-bookings}

\paragraph{Matched service and different agreed rents.}
The Opus--Sol comparison matches world seed, customer request, rental
dates, exact products, quantities, and promised service profiles.
Both bookings must be completed with no omitted promised item or
compensation, and with contracted products delivered unchanged.
Optional items excluded before commitment are outside the service
signature. The sample covers manager-quoted inquiries; initial agreed
dates must also coincide. Table~\ref{tab:matched-rents} reports the
matched transactions.

  \par\medskip\begingroup
  \renewenvironment{table}[1][]{%
    \noindent\begin{minipage}{\columnwidth}\centering\captionsetup{type=table}%
  }{\end{minipage}}%
  \input{tables/generated/matched_rents}  \endgroup\par\medskip

Two seed-42 bookings each rent a table, two comfort chairs, and cookware
for four days. Opus charges 33 and Sol 60. Opus's later notes cite cookware
committed to these bookings when setting aside other fixed-price offers.
Its pricing records repeatedly recommend concessions after rejection,
although higher-price attempts also occur. The match holds delivered
service terms constant; the two shops still have different business
histories. Channel revenues and costs are reported separately in
Appendix~\ref{app:financial-decomposition}.

A complementary Sol seed-73 decision illustrates choosing lower rent for
resource compatibility. On day 78, the customer authorizes three plans
for days 87--90: 39 for a table, lightweight chairs, and cookware; 43 with
comfort chairs instead; and 41 with comfort chairs and a frying pan.
Sol selects the 39 plan as booking 089 rather than the customer's preferred
43 plan. Its note preserves comfort-chair and pan capacity for booking 086
and a possible late return of 087. All three bookings complete without
compensation. The record shows the choice and its stated resource rationale;
it does not assign a measured value to the retained flexibility.

\paragraph{Deliberate exit after economic comparison.}
\label{app:deliberate-exit}
On day 184, DeepSeek seed 42 compares bookings 373 and 387, both due on 185.
Together they exceed usable supply by one cold-weather sleeping bag and
one foam pad; cancelling either removes both shortages. Their rents are
73.50 and 13.20. DeepSeek compares rent, cancellation costs, occupied time,
and the higher compensation and allocation risk of waiting until pickup.
It chooses to retain 373 and cancel 387 a day early.

The executed cancellation pays 7.86 compensation and refunds the 3.30
prepayment. Booking 373 completes for 73.50 net rent and no compensation.
As specified in Appendix~\ref{app:guarantee}, the prepayment refund settles
an existing liability; compensation is the additional exit cost.

Opus seed 73 makes an investment-versus-exit comparison on day 183.
Booking 331 promises a tent and other equipment on days 189--192 for 29,
but its tent overlaps another booking and a date change has been refused.
A second tent costs 89.61, has an estimated two-day delivery, and fits
the recorded 108.26 capital headroom. Those estimates leave time and
capital for procurement six days before pickup.

Opus weighs the acquisition--resale difference, the booking's rent,
possible remaining-season walk-in use, and a 5.18 early-exit charge.
It judges the added investment unattractive and notes that cancellation
also releases the other equipment. The receipt confirms cancellation,
5.18 compensation, and a 7.25 prepayment refund. Its valuation is approximate,
and it remains uncertain about authorization for a cheaper tent substitute.
The example records an explicit comparison of further investment with exit.

\FloatBarrier
\subsubsection{Turning Feedback and Evaluations into Resource Reallocation}
\label{app:capital-comparisons}

\paragraph{Sol: comparing and implementing alternative investments.}
In seed 73, Sol sells a worn lantern on day 41 for 2.40, releasing its
30.90 capital basis the next morning. It compares a sixth comfort chair
at 35.90, a replacement lantern at 30.90, and a three-person pad at 40.
Its review considers recent demand, fleet sizes, delivery times, and
remaining holding costs. Five comfort chairs have supplied 16 units over
37 asset-days for 119.25 rent in the previous thirty days; sixteen
lightweight chairs have supplied 69 units over 112 asset-days for 278.96.

Sol buys the comfort chair on day 42. It rejects adding a seventeenth
lightweight chair despite that fleet's larger total rent: the sixth comfort
chair would enable six-person comfort requests. It judges two remaining
lanterns sufficient and the pad's delivery too late for the nearest need.
On day 45, further rejections of lightweight-chair configurations and
limited comfort supply prompt another adjustment. Sol sells an unused
lightweight chair for 14.30, after checking that fifteen retained units
cover twelve required by upcoming bookings alongside its walk-in
allowance. It buys another comfort chair on 46 after capital is released.

Later, on days 127--128, sparse demand for cold-weather bags and upcoming
bookings using warm-weather bags support selling one cold-weather bag
while retaining two. Repeated group-demand limits and a nine-chair
concurrent requirement support the seating purchases. Sol waits for the
sold bag's 99 capital basis to be released before completing them. These sequences
connect demand comparisons, retained obligations, and executed changes.

\paragraph{Luna: an earlier assessment in a later holding decision.}
In seed 73, Luna evaluates a cheaper comfort chair on day 49. It considers
two units at 35.90 each instead of the existing 102.60 model, discussing
group demand, payback, delivery, and sale of a table with 82 capital basis
to fund the investment. This is a proposed purchase, with procurement
uncertainty also noted.

On day 65, with only 0.98 capital headroom, Luna reviews the higher-cost
chair and supports retention through 37.75 of rental income already earned,
low ongoing costs, and further earning potential. It mentions the cheaper
model as a product it does not currently own, without returning to the
earlier funding proposal.
The purchase of four cheaper chairs occurs on day 116, after sales on 115.
Luna continues other equipment transactions in the interval, including a
mat replacement on days 85--86; the delay concerns this particular
investment rather than all fleet management.

\paragraph{Demand served and reservation business mix.}
\label{app:luna-business-mix}
Over days 49--115, both shops receive 43 walk-in comfort-chair requests
for 141 units. Sol supplies 46 units across 16 served requests and Luna 16
across 7. These are rental occurrences, not distinct assets. They describe
business served under the shops' respective fleets, prices, and limits.

For full-season context on seed 73, the common manager-quoted requests
yield 149 completed bookings for Sol and 148 for Luna, with net collected
rent of 4,181.67 and 3,163.91 and committed asset-days of 967 and 845.
Chair service accounts for 410 and 298 asset-days: Sol's split is 148
lightweight plus 262 comfort, compared with Luna's 256 plus 42.
These reservation totals cover the full season and both chair classes,
whereas the 141-unit comparison concerns comfort-chair walk-ins during
the investment interval.

\subsubsection{Using Available Capacity While Protecting Commitments}
\label{app:capacity-cases}

\paragraph{A protected interval longer than the booking requires.}
Luna seed 73 has a ready three-person pad on day 134 but blocks it from
walk-ins through day 149 for booking 161, whose pickup is on 146.
A local test on day 135 admits an intervening rental scheduled to return on 139,
one week before that pickup. Rent exceeds estimated normal cleaning and
usage costs by 9.76. The candidate passes the one-, three-, and seven-day
turnaround checks in Appendix~\ref{app:relaxation}.

The best qualifying daily candidates span nine products in each Luna run,
in both early and late periods. Table~\ref{tab:relaxation} gives all models'
frequencies and estimated gains, including dates with no improvement.

\paragraph{Remaining opportunities in Sol's configurations.}
For Sol's cookware, the day-80 test in seed 42 adds a rental scheduled
to return on 81, earning \(4.00-1.25-0.186=2.564\) after cleaning and
estimated use. The day-135 test in seed 73 adds one due on 136, earning
\(3.50-1.25-0.186=2.064\). Both pass all three checks.

%% file: tables/generated/matched_rents.tex
\begin{table}[tb]
\centering
\small
\setlength{\tabcolsep}{4pt}
\begin{tabular}{@{}lrrrr@{}}
\toprule
Seed & Pairs & \shortstack{Opus\\rent} & \shortstack{Sol\\rent} & \shortstack{Opus below\\Sol (\%)} \\
\midrule
42 & 44 & 823.90 & 1,183.50 & 30.38 \\
73 & 33 & 480.05 & 763.17 & 37.10 \\
\bottomrule
\end{tabular}
\caption{Total agreed rent (CNY) for matched fully delivered bookings. Opus charges less in 75 of 77 pairs; the percentage difference uses Sol's total rent as its base. Matching criteria are in Appendix~\ref{app:matched-bookings}.}
\label{tab:matched-rents}
\end{table}